\documentclass[11pt]{article}

\usepackage[preprint]{acl}

\usepackage{times}
\usepackage{latexsym}

\usepackage[T1]{fontenc}

\usepackage[utf8]{inputenc}

\usepackage{microtype}

\usepackage{inconsolata}

\usepackage{graphicx}
\usepackage{lipsum}
\usepackage{comment}
\usepackage{multirow}

\usepackage{booktabs}
\usepackage{subcaption}

\usepackage{tcolorbox}
\newcounter{promptno}[section]
\newlength\mystoreparindent
\newenvironment{prompt}[1][]
{
  \setlength{\mystoreparindent}{\the\parindent}
  \setlength{\parindent}{0pt}
  \refstepcounter{promptno}
  \par\medskip
  \noindent
  \begin{tcolorbox}[left=1pt,right=1pt]
  \textsc{{template \small\thesubsection.\thepromptno}}\\
  \small
}{
  \end{tcolorbox}
  \setlength{\parindent}{\mystoreparindent}
  \medskip
}

\title{LLM Judges Can Be Too Generous When There Is No Reference Answer}

\author{Chalamalasetti Kranti \\
  Computational Linguistics, Department of Linguistics,\\
  University of Potsdam, Germany \\
  \texttt{kranti.chalamalasetti@uni-potsdam.de} \\\And
  Sowmya Vajjala \\
  National Research Council, \\
  Ottawa, Canada \\
  \texttt{sowmya.vajjala@nrc-cnrc.gc.ca} \\}

\begin{document}
\maketitle
\begin{abstract}
LLM judges are increasingly being used to evaluate open-ended model responses, often in no-reference settings where a ground-truth answer is unavailable. However, can they reliably assess in such evaluation setups? We explore this question in this paper through a two stage pipeline with a) calibration experiments that assess the judge model's knowledge of the task it is evaluating, and b) sensitivity experiments that assess how the judge model's performance is impacted by the presence and positioning of the reference answer in the prompt. Across experiments covering three languages, we show that the judge models we evaluated tend to over-credit incorrect answers in the absence of a reference answer, and adding reference answer information to the prompt flips the judge model's correct/incorrect decisions by as much as 85\% in some experimental settings. Comparison with a subset of human annotations shows that these reference-driven changes generally align with human judgments. Our results emphasize the need for calibrating the LLM judges with a sample with reference-aware evaluation before using them in reference-free setups reliably, and our methodology provides a blueprint for researchers and practitioners in doing such calibration of LLM judges for other tasks.
\end{abstract}

\section{Introduction}
\label{sec:introduction}
Evaluation of machine generated text has evolved~\citep{reiter2026nlg}, from n-gram-based metrics such as BLEU~\citep{papineni-etal-2002-bleu} and ROUGE~\citep{lin-2004-rouge}, to neural-based metrics such as BERTScore~\citep{zhang2019bertscore}, factuality metrics~\citep{kryscinski-etal-2020-evaluating}, to LLM-based evaluation~\citep{li-etal-2024-leveraging-large}. As the scale of language model output grows, fast, automatic, and accurate evaluation becomes increasingly important. Recent work has leveraged LLMs as evaluators, following the \textit{LLM-as-a-Judge} paradigm, across a wide range of tasks~\citep{gu2024survey, li2024llms, li-etal-2025-generation}. Despite this widespread adoption, researchers have uncovered various shortcomings of LLMs as judges \citep[e.g.,][]{doddapaneni-etal-2024-finding,razavi2025benchmarking,moon-etal-2026-dont}. While these studies primarily study the various aspects around the reliability of LLM-judge decisions, the question of the judge model's ability to do the evaluation task or its sensitivity to the presence/absence of ground-truth information in the evaluation prompt was not studied much in the past research.

\begin{figure}[t]
  \includegraphics[width=\columnwidth]{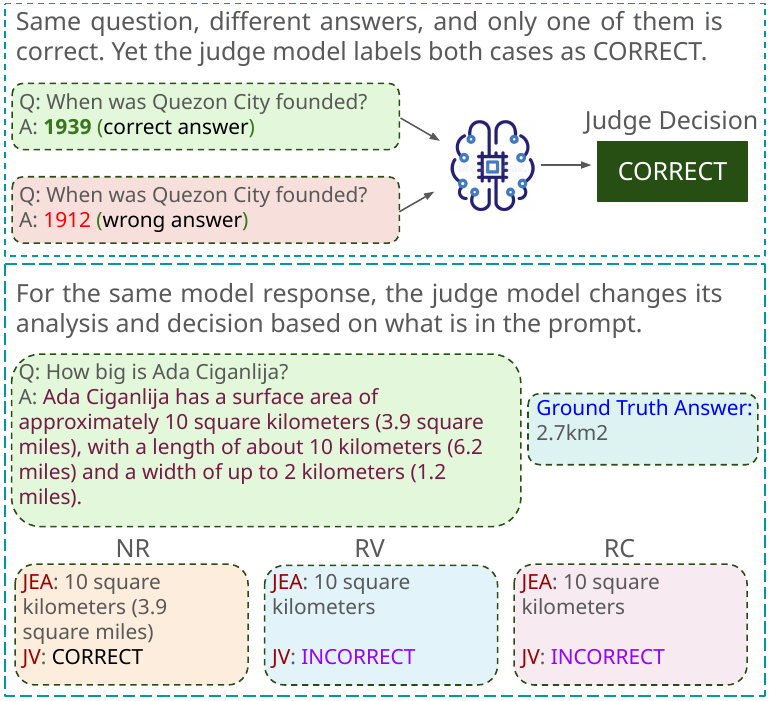}
  \caption{LLM-judge verdicts can be over-credited in no-reference evaluation. Top: without a reference, the judge marks both a correct answer and an incorrect answer as CORRECT. Bottom: the judge marks the answer as CORRECT in NR, but changes the verdict to INCORRECT in RV and RC. NR: no reference; RV: reference visible; RC: reference comparison; JEA: judge-extracted answer; JV: judge verdict.}
  \label{fig:firspagefig}
\end{figure}

Many real-world tasks that use LLM judges are evaluated in reference-free settings because ground-truth information is not always readily available. In this setting, the judge must evaluate correctness based on its own knowledge of the domain, language, and task. When this knowledge is insufficient, the judge may accept incorrect answers, which can inflate reported correctness scores. Figure~\ref{fig:firspagefig} shows an example where the judge marks both correct and incorrect answers as correct in a no-reference setting, but changes its decision when the ground truth is present. \textit{So, how do we identify whether an LLM judge can reliably do reference-free evaluation for a given task?} We explore this question in this paper in an controlled open-ended, multilingual question answering setup through a novel two stage pipeline: (a) calibration experiments, which evaluate whether the LLM judge has a good understanding of the task it is evaluating, and (b) sensitivity experiments, which assess whether the LLM judge can be reliably used in a reference-free evaluation scenario. These experiments allow us to systematically analyze whether evaluation failures of LLM judges are caused by the limitations of the judge model's capability for that task  itself or because of its sensitivity to the presence of a reference answer while judging. In practice, this two stage pipeline can also serve as a blueprint to develop similar evaluation pipelines for comparing and choosing the right LLM judges for other tasks. 

\section{Related Work}
\label{sec:relatedwork}
\paragraph{Large Language Models as Evaluators}
Recent work has increasingly used LLMs as automatic evaluators for tasks such as textual coherence~\citep{barbosa-campelo-2024-llms}, mathematical reasoning \citep{li-etal-2025-exploring-reliability, stephan-etal-2025-calculation}, code generation and evaluation~\citep{zhao-etal-2025-codejudge, 10.1145/3728963, moon-etal-2026-dont}, biomedical relation extraction~\citep{laskar-etal-2025-improving}, automatic answer grading~\citep{RODRIGUES2025100428, su-etal-2025-essayjudge, DBLP:conf/ecai/ZhuHCCLM25}, text summarization~\citep{NEURIPS2025_829e8f32} and open-ended question evaluation~\citep{3666122.3668142, kamalloo-etal-2023-evaluating, wei2026qurl} among others. Research on fine-tuned evaluator models~\citep{ICLR2025_7f8f7313, huang-etal-2025-empirical} and benchmarks for evaluating LLM judge behavior also exist ~\citep{ICLR2024_afc8b034, ICLR2025_9e720fce, xu-etal-2025-context}. While LLM judges offer a practical alternative to human evaluation, their reliability depends on prompt design, task framing, and the information made available during evaluation. For example, ~\citet{DBLP:journals/corr/abs-2601-07506} shows that judges fail to follow a provided reference when it conflicts with their parametric knowledge. In this paper, we focus on the changes in judge behavior in the presence/absence of reference answers in a multilingual experimental setup.

\paragraph{Biases in LLM Judge Evaluation} Prior work has studied several limitations of LLM judges~\citep{DBLP:journals/corr/abs-2411-15594, li-etal-2025-generation}, including prompt sensitivity~\citep{echterhoff-etal-2024-cognitive, thakur-etal-2025-judging}, position bias~\citep{wang-etal-2024-large-language-models-fair, li-etal-2024-split, jiao2024enhancing, ICLR2025_fdca08d3, shi-etal-2025-judging}, verbosity bias~\citep{hongli-etal-2024-mitigating, jiao2024enhancing, Alvarez-Arenas2026.06.15.26355670}, gender bias, authority bias~\citep{chen-etal-2024-humans}, self-preference bias~\citep{chen-etal-2025-beyond} and inconsistencies across evaluation rubrics~\citep{doddapaneni-etal-2024-finding, lee-etal-2025-checkeval, wu-aji-2025-style, siro-etal-2026-learning}. Our work instead examines how reference information in the prompt affects judge behavior, and whether reference visibility and comparison framing, change verdicts across languages. To our knowledge, these concerns were not addressed sufficiently in the previous work, especially in a multilingual context.

\section{Methodology}
\label{sec:methodology}
Our objective is to study the reliability of LLM judges in reference-free evaluation setups, focusing on both their ability to do the task as well as their sensitivity to the presence of a reference answer in the prompt. For this purpose, we conduct a controlled study in which an LLM judge evaluate model-generated answers to open-ended questions and assigns a binary verdict of correct or incorrect. 

\begin{figure}[htb!]
  \includegraphics[width=\columnwidth]{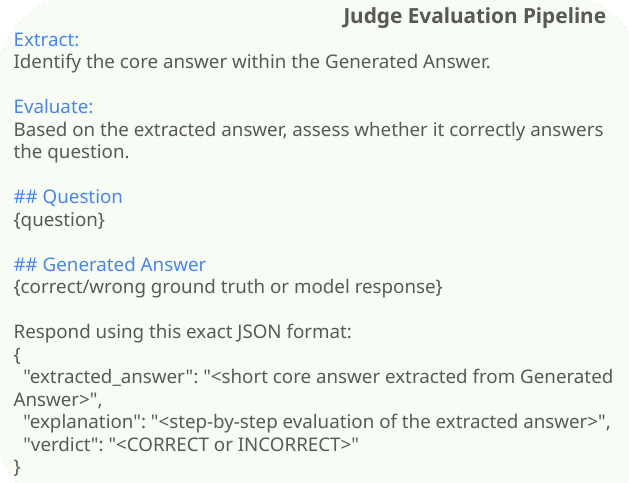}
  \caption{LLM-judge evaluation pipeline used across calibration and sensitivity experiments. The judge extracts the core answer from the model response and then assigns a correctness verdict and responds in JSON format.}
  \label{fig:evalpipeline}
  \vspace{-0.3cm}
\end{figure}

The general experimental pipeline (see Figure~\ref{fig:evalpipeline}) has two components: First, the LLM judge is asked to \textit{extract the answer} from the generated model response, since open-ended responses are often elaborate. Then, the judge checks whether the \textit{extracted answer is correct for the given the question}, and assigns a decision of \textit{CORRECT} or \textit{INCORRECT}. The final output of the judge is a JSON object with three fields: \texttt{extracted\_answer}, \texttt{explanation}, and \texttt{verdict}. The \texttt{extracted\_answer} field has the answer extracted by the judge from the model response, the \texttt{verdict} field gives the final binary decision, and the \texttt{explanation} field provides the rationale for the decision. This approach allows us to evaluate what the judge identifies as the answer, how this varies with context, and how the final decision is made. We design two categories of experiments to study the behavior of LLM judges with this pipeline: calibration experiments and sensitivity experiments. 

\begin{figure}[t]
  \includegraphics[width=\columnwidth]{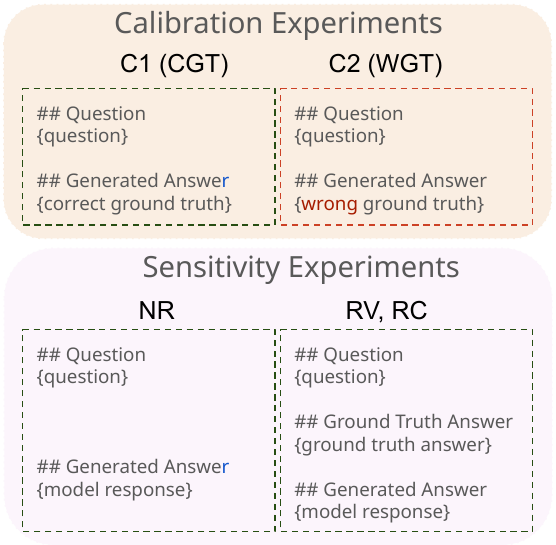}
  \caption{Overview of calibration and sensitivity experiment settings. Calibration varies answer correctness, while sensitivity varies the availability and framing of the reference answer.}
  \label{fig:calibsensoverview}
  \vspace{-0.3cm}
\end{figure}

\subsection{Calibration Experiments}
\label{subsec:calibrationmethodology}
These experiments test whether the judge model is knowledgeable about the task. A question and a possible answer are provided and the judge has to evaluate whether the answer is correct. We run this experiment in two settings: one in which the answer provided corresponds to the correct ground-truth answer (C1), and the other in which it corresponds to an incorrect ground-truth answer (C2). In both settings, the answer is labeled as \textit{Generated Answer} (see Figure~\ref{fig:calibsensoverview}), so the judge must rely on its own reasoning rather than an explicit label \footnote{The prompt can be found in Figure~\ref{fig:llmjudge_calibration_prompt} in the Appendix}. To avoid making incorrect cases easy to detect, we construct the incorrect ground truth answers from other existing question-answer pairs in the datasets. Note that in this setting, we aim to only calibrate the judge model for the task, and it does not require any generator models. 

\subsection{Sensitivity Experiments}
\label{subsec:sensitivitymethodology}
The sensitivity experiments aim to estimate the suitability of the judge to do reference-free evaluation by measuring how the presence and positioning of a reference answer changes a judge's evaluation of generator model's responses, through three experimental settings: No-Reference (NR), Reference-Visible (RV), Reference-Compared (RC)\footnote{See Figures~\ref{fig:llmjudge_sensititivy_prompt_nr}--~\ref{fig:llmjudge_sensititivy_prompt_rcp} in the Appendix for the full prompts used in the evaluation.}.

In the no-reference setting (NR), the judge evaluates whether the model-generated response correctly answers the question without access to any reference. This experiment differs from the calibration experiments as the answer being evaluated here is the model-generated response. In the reference-visible settings (RV), the reference answer is present in the prompt but is not explicitly used for comparison. In the reference-comparison setting (RC), the prompt explicitly instructs the judge to compare the extracted answer with the reference answer before assigning a decision. 

\paragraph{Evaluation with Decision Flips: }
\label{subsec:decisionflipsmethodology}
For all the experiments, we measure the decision flips i.e., changes in the judge model's evaluations (from \textit{CORRECT} to \textit{INCORRECT}, or vice versa)  between the experimental setups. In the calibration experiments, we would expect that the difference between C1 and C2 is high if a judge model has a good understanding of the task. For all sensitivity experiments, we treat the no-reference (NR) setting as the baseline and analyze decision flips across two types of transitions: NR to RV, which captures the effect of making the reference visible; RV to RC, which captures the effect of explicitly requiring comparison with the reference. If a judge model is good, we expect the amount of such decision flips should be as low as possible. 

\section{Experimental Setup}
\label{sec:expsetup}
We evaluate two datasets - one covering three languages (English, Arabic and Telugu), and the other covering one of these three (Telugu). All examples are evaluated using responses generated by four LLMs, and each response evaluated by three judge models, resulting in twelve judge evaluations per dataset-language pair. 

\paragraph{Datasets: } We run our experiments with two data sources to examine whether the findings generalize across datasets and languages. TyDi QA \cite{clark2020tydi}~\footnote{Downloaded from \url{https://huggingface.co/datasets/copenlu/answerable_tydiqa}} is a question-answering dataset covering 11 languages. Each language subset has training and validation splits and we use the validation splits for Telugu ($1338$), Arabic ($1902$), and English ($990$). We further process these data to exclude entries with unanswerable questions, for which the ground truth is null. This results in a final count of $669$ questions in Telugu, $951$ questions in Arabic, and $445$ questions in English. The languages come from different language families, and each uses a different writing system.  

\begin{figure*}[htbp]
    \centering
    \begin{subfigure}{0.5\textwidth}
        \centering
        \includegraphics[width=\textwidth]{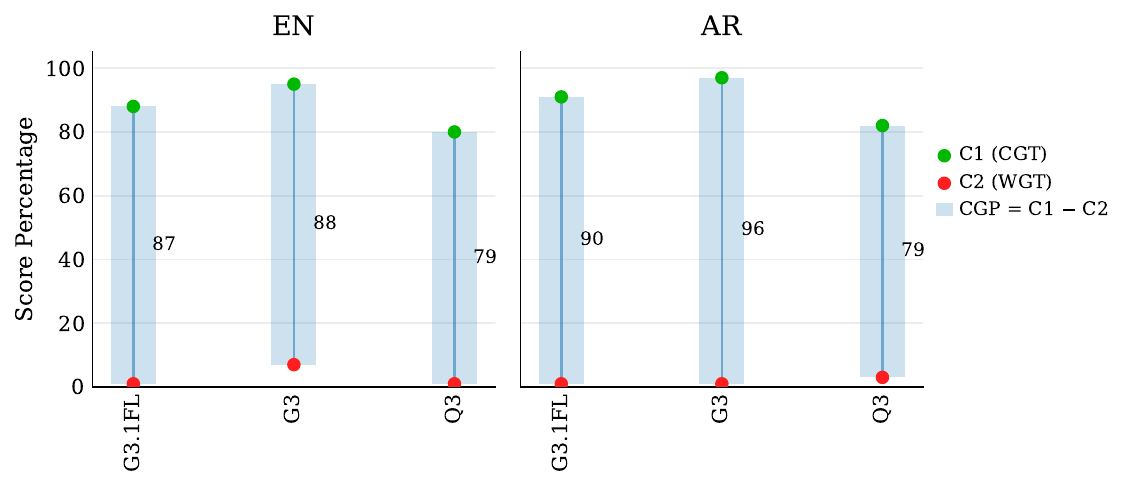}
        \caption{Calibration results for English and Arabic in TyDi QA}
        \label{fig:en_ar_gap}
    \end{subfigure}
    \hfill
    \begin{subfigure}{0.4\textwidth}
        \centering
        \includegraphics[width=\textwidth]{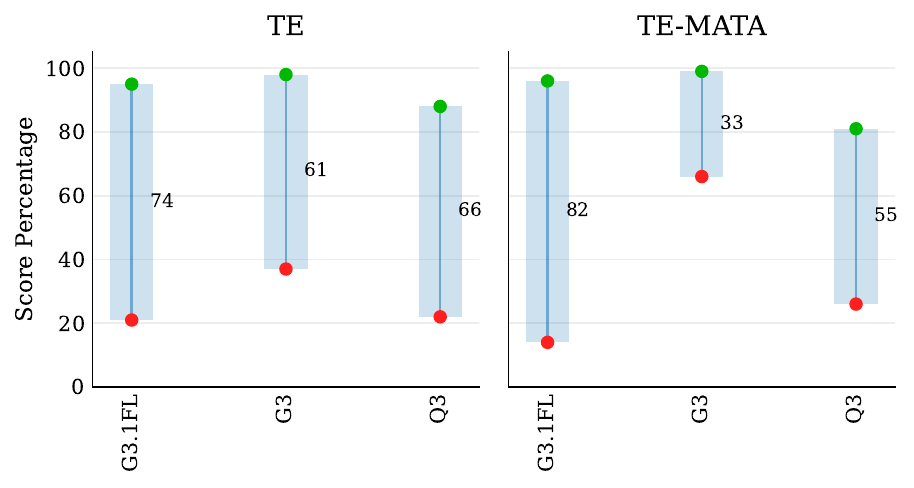}
        \caption{Calibration results for Telugu in TyDi QA and MATA}
        \label{fig:te_te_gap}
    \end{subfigure}

    \caption{Calibration results across languages and datasets. C1 measures how often judges accept correct ground-truth answers, while C2 measures how often they accept incorrect answers; All judges over-credit incorrect answers (C2 > 0) for Telugu; CGP is the difference between C1 and C2, with higher values indicating better separation; G3.1FL: Gemini3.1-Flash-Lite-Preview; G3: Gemma3-27B; Q3: Qwen3-32B; }
    \label{fig:calibexperiments}
    \vspace{-0.3cm}
\end{figure*} 

The second dataset is  \textsc{MATA} \cite{kranti-etal-2026-mata}~\footnote{\url{https://huggingface.co/datasets/TeluguLLMResearch/MATA}}, which is a recent dataset for evaluating the Telugu language capabilities of LLMs, and covering seven categories of questions including factual knowledge and various forms of linguistic reasoning, grammar and vocabulary as well. In comparison, TyDiQA is primarily focused on factual questions. We use only the open-ended questions from \textsc{MATA} (540 questions) in our evaluation. Furthermore, we use the question category information available in the dataset to select a wrong-answer from the same category for the calibration experiments.  

\paragraph{Model Selection} We use four models to generate responses to the input questions: two general models Gemini 3.1-Pro \cite{gemini31pro} and Qwen3-32B \cite{qwen3} and one language-optimized model for each non-English language, Sarvam-105B \cite{sarvam} for Telugu and Fanar-C-2-27B \cite{fanarteam2025} for Arabic. For judge evaluation, we choose three judge models: two open-weight models, Qwen3-32B and Gemma3-27B \cite{gemmateam2025}, and one closed model, Gemini-3.1-FlashLite-Preview (Now available as Gemini-3.1-Flash-Lite~\citep{gemini31flashlite}). This selection allows us to examine the effect of model type on evaluation behavior. All evaluations are run using the Inspect~\footnote{\url{https://inspect.aisi.org.uk/}} framework, with prompts provided in the Appendix~\ref{subsec:appendix-prompt-templates}. For model access, Sarvam-105B is accessed through the Sarvam API\footnote{\url{https://www.sarvam.ai/}}, Fanar-C-2-27B is accessed through the FanAR API\footnote{\url{https://api.fanar.qa/}}, and the remaining models are accessed through OpenRouter~\footnote{\url{https://openrouter.ai/}} API. All models are evaluated in a zero-shot setting with \textit{temperature=0}. We deliberately include overlap between response generator and judge models, either through the same model, Qwen3-32B, or through models from the same family, Gemini/Gemma. This design allows us to study whether judge behavior is affected by self-model or same-family evaluation.  

\paragraph{Human Evaluation} We conduct human evaluation on a sample of responses from the sensitivity experiments with the \textsc{MATA} dataset. The authors, 
who are native Telugu speakers, evaluate $400$ model responses from this dataset, sampled from two question subcategories: Factual Knowledge and Reasoning. These human generated correctness labels for model responses are compared against the judge verdicts in the NR, RV, and RC sensitivity experiments. 

\section{Results}
\label{sec:results}
We first present the results of the calibration experiments that assess the judge's understanding of the task, and then discuss in detail the various aspects of the sensitivity experiments including a human annotation study.\footnote{Some examples of LLM judge responses can be found in Appendix~\ref{subsec:appendix-qualitative-analysis}, which show that judges may follow, override, or reinterpret reference information depending on how it is presented in the prompt.}

\subsection{Calibration of LLM Judges}
\label{subsec:calibresults}
The calibration experiments (Section~\ref{subsec:calibrationmethodology}) aim to answer the question \textit{How well can LLM judges distinguish between correct and incorrect answers?} Figure~\ref{fig:calibexperiments} presents the results. We have two experimental settings: C1 (where the provided answer is the correct ground-truth answer) and C2 (where the provided answer is an incorrect answer). Ideally, we would want the judge performance in C1 to be closer to 100\% and in C2 to be closer to 0, with a large difference between C1 and C2. For English and Arabic, we observe that this is indeed the case, as shown in Figure~\ref{fig:en_ar_gap}. However, for Telugu, we observe C2 scores as high as 60\% for one judge model (Gemma3-27B on \textsc{MATA} in Figure~\ref{fig:te_te_gap}), indicating a tendency of that judge model to over-credit incorrect answers rather than flagging them as wrong in lower-resource settings. It could mean that while the chosen judge models appear reasonably calibrated for the question answering task in English and Arabic, they may not be a good choice in a relatively low-resource language like Telugu. In terms of practical relevance, performing such a calibration experiment with different LLM judges on a smaller evaluation set which has gold standard reference answers can offer a good method to estimate whether a given LLM judge has a sound understanding of the task it is evaluating. \footnote{We tested a variation in which the judge is allowed to choose among: \textit{CORRECT}, \textit{INCORRECT}, and \textit{I don't know}. None of the judges used \textit{I don't know} in their verdicts, suggesting that the models made explicit correctness decisions.}

\begin{table*}[htb!]
\centering
\tiny
\setlength{\tabcolsep}{2.5pt}
\resizebox{\textwidth}{!}{%
\begin{tabular}{llccc|ccc|ccc|ccc}
\toprule
\multirow{2}{*}{JM} & \multirow{2}{*}{RM}
& \multicolumn{3}{c|}{EN}
& \multicolumn{3}{c|}{AR}
& \multicolumn{3}{c|}{TE}
& \multicolumn{3}{c}{TE-MATA} \\
\cmidrule(lr){3-5}
\cmidrule(lr){6-8}
\cmidrule(lr){9-11}
\cmidrule(lr){12-14}
& & NR & RV ($\Delta$) & RC ($\Delta$)
  & NR & RV ($\Delta$) & RC ($\Delta$)
  & NR & RV ($\Delta$) & RC ($\Delta$)
  & NR & RV ($\Delta$) & RC ($\Delta$) \\
\midrule

G3.1FL & G3.1P
& 0.99 & 0.87 (0.12) & 0.78 (0.09)
& 0.99 & 0.92 (0.07) & 0.88 (0.06)
& 0.95 & 0.34 (0.61) & 0.32 (0.02)
& 0.96 & 0.91 (0.09) & 0.90 (0.02) \\

       & Q3
& 0.90 & 0.76 (0.16) & 0.67 (0.09)
& 0.82 & 0.70 (\textbf{0.14}) & 0.65 (0.06)
& 0.83 & 0.14 (0.69) & 0.13 (0.02)
& 0.46 & 0.31 (0.17) & 0.31 (0.02) \\

       & SM
& 0.86 & 0.72 (\textbf{0.18}) & 0.68 (0.08)
& 0.89 & 0.89 (0.00)  & 0.66 (0.24)
& 0.90 & 0.19 (\textbf{0.71}) & 0.16 (0.03)
& 0.60 & 0.30 (\textbf{0.33}) & 0.29 (0.03) \\

       & FA
& 0.95 & 0.79 (\textbf{0.18}) & 0.70 (0.10)
& 0.89 & 0.80 (0.11)  & 0.73 (0.06)
& 0.89 & 0.19 (\textbf{0.71}) & 0.17 (0.02)
& 0.46 & 0.29 (0.18) & 0.27 (0.03) \\

\midrule

G3 & G3.1P
& 1.00 & 0.82 (0.17) & 0.78 (0.06)
& 0.99 & 0.89 (0.11) & 0.87 (0.03)
& 1.00 & 0.40 (0.60) & 0.40 (0.01)
& 0.99 & 0.94 (0.07) & 0.94 (0.01) \\

   & Q3
& 0.99 & 0.76 (0.12) & 0.69 (0.04)
& 0.99 & 0.74 (\textbf{0.25}) & 0.69 (0.06)
& 0.98 & 0.21 (\textbf{0.78}) & 0.17 (0.05)
& 0.97 & 0.48 (\textbf{0.50}) & 0.48 (0.09) \\

   & SM
& 0.99 & 0.75 (\textbf{0.24}) & 0.70 (0.07)
& 0.99 & 0.79 (0.20) & 0.76 (0.04)
& 0.99 & 0.35 (0.65) & 0.32 (0.07)
& 0.97 & 0.55 (0.44) & 0.54 (0.04) \\

   & FA
& 0.99 & 0.82 (0.18) & 0.74 (0.09)
& 0.98 & 0.80 (0.18) & 0.78 (0.04)
& 0.98 & 0.36 (0.61) & 0.30 (0.12)
& 0.93 & 0.44 (0.50) & 0.42 (0.07) \\

\midrule

Q3 & G3.1P
& 0.93 & 0.69 (0.29) & 0.61 (0.13)
& 0.91 & 0.59 (0.37) & 0.54 (0.12)
& 0.92 & 0.30 (0.69) & 0.28 (0.08)
& 0.83 & 0.81 (0.28) & 0.82 (0.08) \\

   & Q3
& 0.97 & 0.62 (\textbf{0.37}) & 0.52 (0.12)
& 0.94 & 0.08 (\textbf{0.60}) & 0.08 (0.03)
& 0.96 & 0.11 (\textbf{0.85}) & 0.11 (0.04)
& 0.83 & 0.32 (\textbf{0.54}) & 0.31 (0.08) \\

   & SM
& 0.90 & 0.57 (0.36) & 0.52 (0.11)
& 0.89 & 0.57 (0.36) & 0.54 (0.12)
& 0.92 & 0.18 (0.77) & 0.15 (0.04)
& 0.75 & 0.40 (0.48) & 0.40 (0.09) \\

   & FA
& 0.94 & 0.64 (0.35) & 0.58 (0.10)
& 0.91 & 0.65 (0.30) & 0.59 (0.11)
& 0.92 & 0.14 (0.82) & 0.13 (0.02)
& 0.63 & 0.26 (0.43) & 0.25 (0.08) \\

\bottomrule
\end{tabular}
}
\caption{Correctness score changes across EN, AR, TE languages in Ty Di QA dataset and TE language in MATA datasets. JM: Judge Model; RM: Response Model; NR: No Reference; RV: Reference Visible; RC: Reference Comparison; G3.1FL: Gemini3.1-Flash-Lite-Preview; G3: Gemma3-27B; Q3: Qwen3-32B; G3.1P: Gemini3.1-Pro; SM: Sarvam-105B; FA: Fanar-C-2-27B; Delta is the flip rate and is computed as the average fraction of total modifications for a given configuration.}
\label{tab:correctnessstats}
\vspace{-0.3cm}
\end{table*}

\subsection{Sensitivity Analysis of LLM Judges}
\label{subsec:sensitivityresults}
We now turn to sensitivity experiments, where our goal is to address the question: \textit{How does the presence and positioning of the reference answer affects the LLM judge decisions?}. Note that while the calibration experiment helped us understand whether a judge model is good at the question answering task, the sensitivity experiments address the question of whether we can reliably do reference free evaluations with that model.  

Table~\ref{tab:correctnessstats} shows the results for the three experimental settings (see Section~\ref{subsec:sensitivitymethodology} for details): No Reference (NR), Reference Visible (RV), and Reference Comparision (RC). \footnote{For both RV and RC, we additionally test a prompt variant based on the position of the reference answer: the default RV (Question, Reference Answer, Generated Answer) versus RVP (Question, Generated Answer, Reference Answer), and similarly, RC and RCP. The results for the position-variant settings (RVP/RCP) are in Appendix~\ref{subsubsec:appendix-positionexp}, and the findings are similar to RV/RC.} We report correctness scores for NR, RV, and RC, along with the decision flip rates across settings. The RV flip rate (delta) is the proportion of questions for which the judge verdict changes from NR to RV, while the RC flip rate is the proportion of questions for which the verdict changes from RV to RC. 

Across all three languages, correctness scores are highest in the NR setting and decrease when the reference is added in RV. Scores decrease further in RC, where the judge is explicitly asked to compare the model response against the ground-truth answer. This pattern suggests that judges may over-credit responses in the NR setting and retract some of these verdicts when reference information is available. In addition, the largest changes in flip rates occur from NR to RV rather than from RV to RC. The NR-to-RV flips range from $0.09$ to $0.85$, while the RV-to-RC flips range from $0.01$ to $0.24$. This suggests that judges use visible reference information and change their verdicts even when the prompt does not explicitly ask them to compare against the reference. The effect is observed across English, Arabic, and Telugu, showing that reference-driven changes in judge decisions are not language specific. 

Across judges, Gemini3.1-FlashLite-Preview shows smaller flip rates from NR to RC for English and Arabic in TyDi QA, with maximum flips of $0.18$ and $0.14$, respectively. But, the flips for this model are larger for Telugu, with the highest at $0.71$ for the responses from Sarvam-105B model on TyDi QA dataset. A similar pattern is observed for Gemma3-27B judge, with maximum flips of $0.24$, $0.25$, $0.50$, and $0.78$ for English, Arabic, Telugu-MATA, and Telugu-TyDi QA, respectively. Qwen3-32B shows the largest flip rates among the judges across languages and response models, with the highest flips observed for Qwen3-32B-generated responses. For English, the overall NR-to-RC flips range from $0.29$ to $0.37$. The effect increases for Arabic, where the flips range from $0.36$ to $0.60$, and is highest for Telugu, where it ranges from $0.69$ to $0.85$. These high NR scores and larger NR-to-RC flip rates suggest that NR scores are highly inflated for Telugu across datasets, especially when Qwen3-32B is used as the judge. This again highlights that reference visibility can have a stronger effect in lower-resource settings.

\begin{figure*}[htbp]
    \centering
    \begin{subfigure}{0.55\textwidth}
        \centering
        \includegraphics[width=\textwidth]{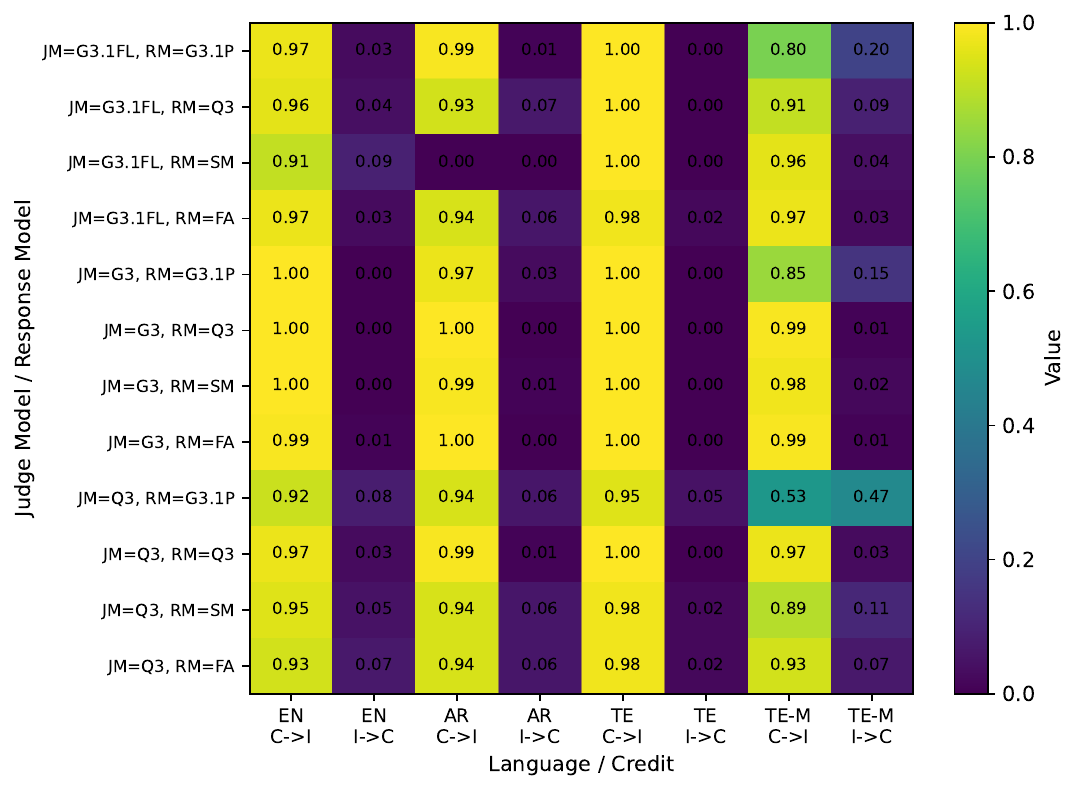}
        \caption{NR-RV Flips}
        \label{fig:heatmap_nr_rv}
    \end{subfigure}
    \hfill
    \begin{subfigure}{0.42\textwidth}
        \centering
        \includegraphics[width=\textwidth]{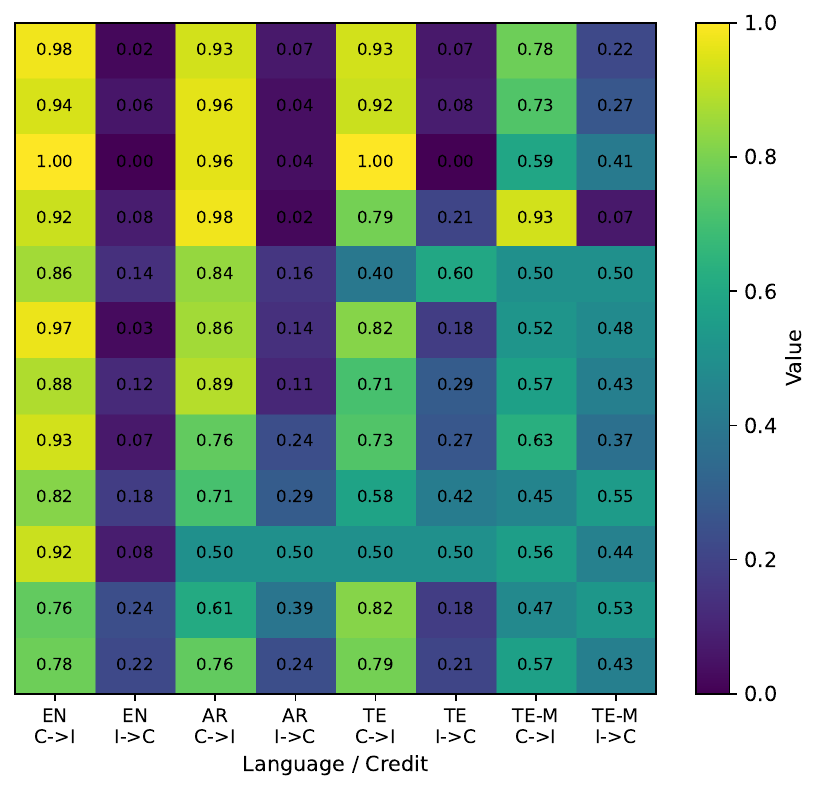}
        \caption{RV-RC Flips}
        \label{fig:heatmap_rv_rc}
    \end{subfigure}

    \caption{Direction of flips across EN, AR, TE, and TE-M (MATA dataset) languages. JM: Judge Model; RM: Response Model; NR-RV: No Reference to Reference Visible; RV-RC: Reference Visible to Reference Comparison; G3.1FL: Gemini3.1-Flash-Lite-Preview; G3: Gemma3-27B; Q3: Qwen3-32B; G3.1P: Gemini3.1Pro; SM: Sarvam-105B; FA: Fanar-C-2-27B.}
    \label{fig:flipsdirection}
    \vspace{-0.3cm}
\end{figure*}

These sensitivity experiments show that the no reference setting produces higher correctness scores across languages, but these scores decrease once reference information is introduced. While this drop is somewhat lesser for English and Arabic, it is strongest for Telugu across both the datasets, suggesting that a reference-free evaluation can potentially result in an inflated amount of correct responses, especially in lower-resource settings.

\paragraph{How do judge verdicts flip across settings?}
\label{subsec:flipdirection}
The sensitivity experiments show that overall judge correctness scores drop across settings, but these scores do not show what kinds of decision changes trigger these drops. To understand this, we examine the decision flips where a judge changes its verdict from \textit{CORRECT} (C) to \textit{INCORRECT} (I) or vice versa across settings more closely. We refer to C->I flip, where the judge withdraws credit it previously granted as \textsc{over credit}, and an I->C flip, where it grants credit it previously rejected as \textsc{under credit} respectively.

Figure~\ref{fig:flipsdirection} shows the results. Across languages and datasets, most NR-to-RV flips are C->I flips. This aligns with the earlier observations in Sections~\ref{subsec:calibresults} and~\ref{subsec:sensitivityresults} that the correctness scores are inflated in the NR setting. The only exception is Qwen3-32B judging Gemini3.1-Pro responses on the MATA dataset, where the flips are more evenly split between C->I and I->C flips, with proportions of $0.53$ and $0.47$, respectively. RV-to-RC flips show a different pattern. For English, and for Gemini3.1-FlashLite Preview as the judge, most RV-to-RC flips are C->I flips. However, for the other languages and judges, I->C accounts for $30\%$ to $70\%$ of RV-to-RC flips. 

This suggests that explicit comparison with the reference does not only make judges more restrictive. In some cases, it also leads judges to accept answers that were rejected especially for low-resource languages like Telugu. \textit{Thus, reference framing changes both the frequency and direction of judge decisions}.

\paragraph{Are the extracted answers same across settings?}
\label{subsec:extractedanswer}

\begin{figure*}[htb!]
    \centering
    \begin{subfigure}{0.49\textwidth}
        \centering
        \includegraphics[width=\textwidth]{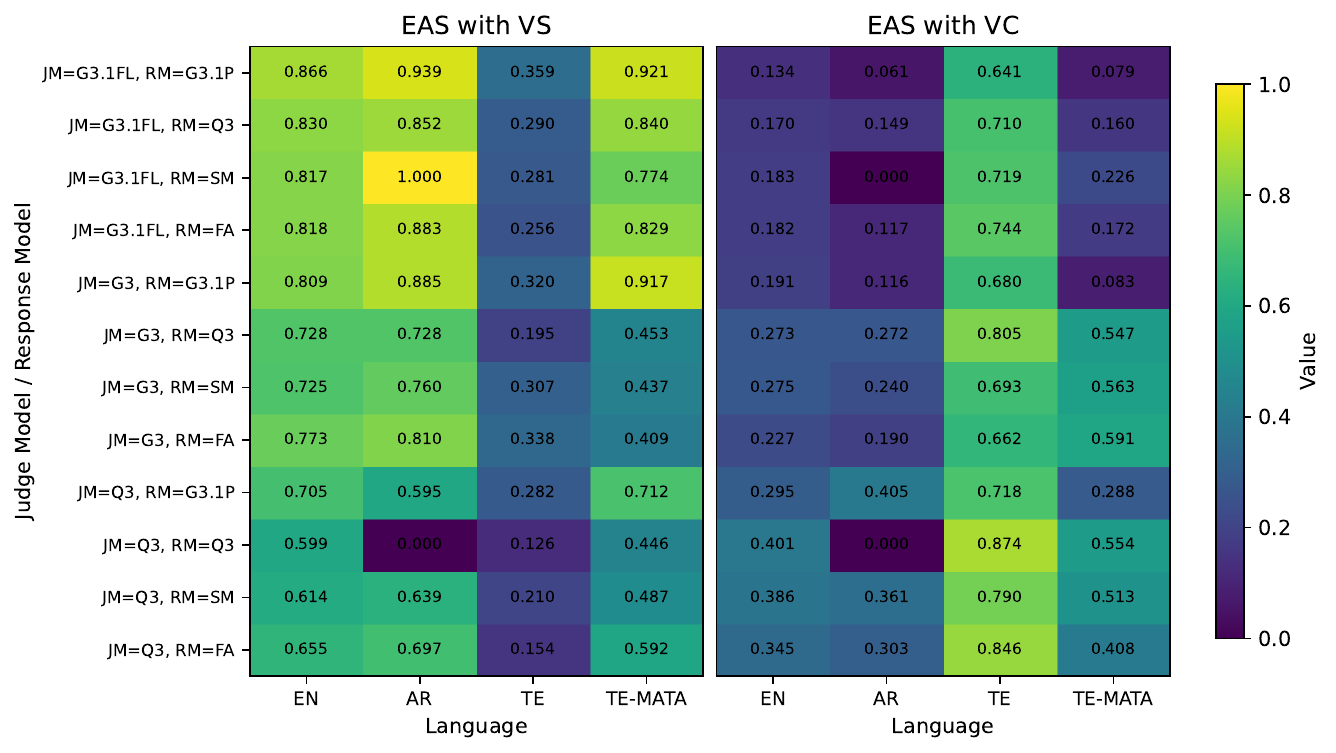}
        \caption{EAS Flips}
        \label{fig:heatmap_eas_vs_vc}
    \end{subfigure}
    \hfill
    \begin{subfigure}{0.49\textwidth}
        \centering
        \includegraphics[width=\textwidth]{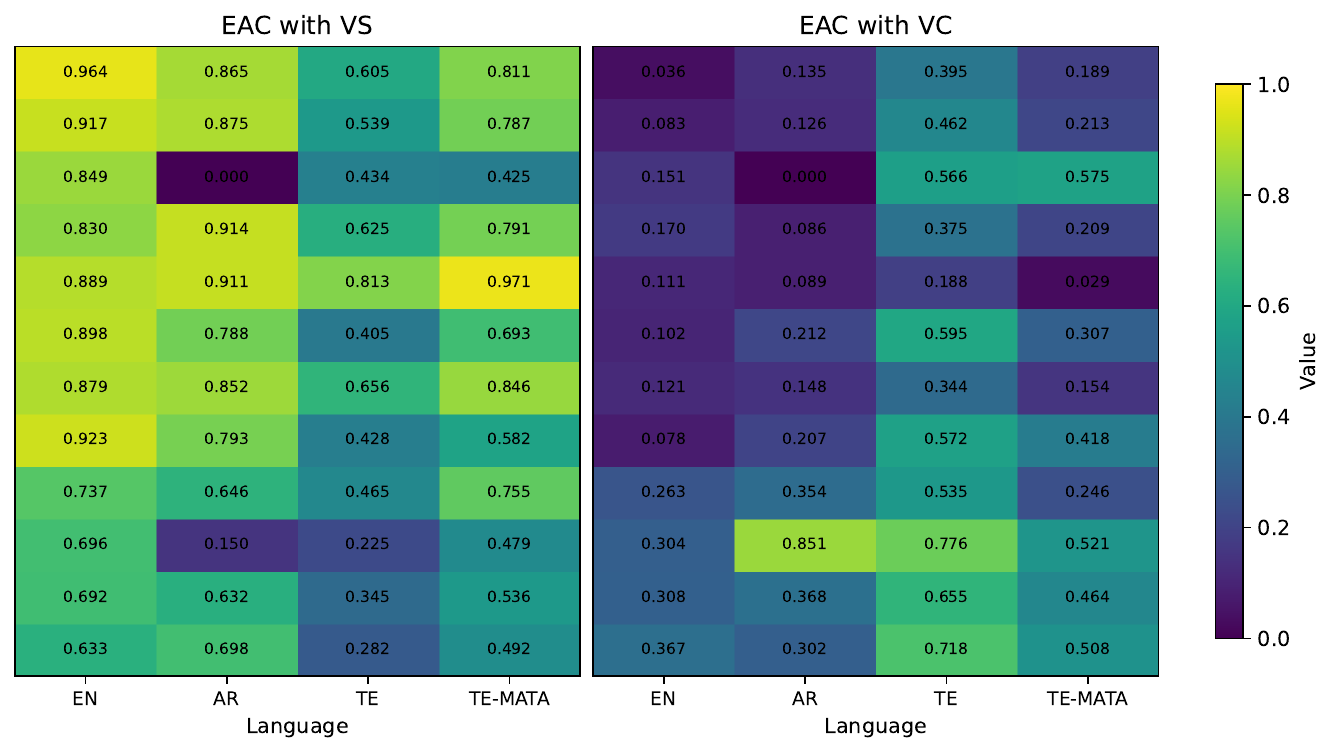}
        \caption{EAC Flips}
        \label{fig:heatmap_eac_vs_vc}
    \end{subfigure}

    \caption{Extracted Answer vs Verdict Stability across EN, AR, TE, and TE-MATA languages. JM: Judge Model; RM: Response Model; EAS: Extracted answer same; EAC: Extracted answer changed; VS: Verdict Same; VC: Verdict Changed.
    }
    \label{fig:eavsverdict}
    \vspace{-0.3cm}
\end{figure*}

We now know that the LLM judges flip their verdicts based on the availability of a reference answer. However, why/how does this happen? One potential spot to look at is what the judge is seeing as the right answer in the entire model generated response. \textit{Is the judge extracting a different answer from the model generated response based on the presence or absence of a reference answer in the prompt?} If so, that can explain the flips to some extent. 

To understand if that is the case, we analyze how the \texttt{extracted\_answer} (See Figure~\ref{fig:evalpipeline})) changes when reference is provided in the prompt, and whether this is accompanied by the changes in judge verdicts. We do an exact match of the extracted answer across settings, and analyze how the judge verdicts change. Figure~\ref{fig:eavsverdict} shows the results of this analysis.\footnote{Table~\ref{tab:easignals-all-judges} in the Appendix show the detailed results across languages, datasets, judge models, and response models.} There are two cases of verdict change that may arise because of the presence of a reference answer: a) the judge may change verdict because the extracted answer changed; b) it can extract the same answer but evaluate it differently in the presence of a reference answer, resulting in a verdict change. 

We observe that the extracted-answer changes are higher for NR-to-RV than for RV-to-RC. For English and Arabic, these extracted-answer changes lead to verdict changes in about half of the cases, while the verdict remains the same in the other half. This aligns with the small to moderate correctness drops observed in Section~\ref{subsec:sensitivityresults}. For Telugu, however, verdict changes occur more often even when the extracted answer remains the same. \textit{This confirms that the reference information drives how the same answer is evaluated rather than which answer is identified, with a larger effect in the low-resource settings}. 

\subsection{Comparing with Human Annotations}
\label{subsubsec:humanannotations}
While the sensitivity experiments show that judges become more restrictive in RC compared to NR, this does not necessarily imply better evaluation. We therefore did a human annotation study to understand whether the stricter verdicts by LLM judges align more closely with human judgments. We chose a sample from the \textsc{MATA} dataset consisting of Telugu questions from two categories: Factual Knowledge and Other Reasoning. We choose these categories because they show higher flip rates in our experiments. The sample consists of $50$ open-ended questions for each category ($100$ in total) and model outputs from four response models ($400$). The human agreement scores (Cohen's Kappa) are in the range of $0.96$-$0.99$ and the details are available in Section~\ref{subsubsec:appendix-humaneval} in the Appendix. 

\begin{table}[htb!]
\centering
\scriptsize
\begin{tabular}{ccccccc}
\hline
\multirow{2}{*}{JM} & \multicolumn{2}{c}{NR} & \multicolumn{2}{c}{RV} & \multicolumn{2}{c}{RC} \\
       & H1    & H2   & H1    & H2   & H1    & H2   \\ \hline
G3.1FL & 0.74 & 0.74 & \textbf{0.96} & \textbf{0.96} & \textbf{0.96} & \textbf{0.96}   \\
G3   & 0.34 & 0.33 & 0.85 & 0.85 & 0.86 & 0.86      \\
Q3   & 0.42 & 0.42 & 0.90 & 0.90 & 0.89 & 0.89      \\ \hline
\end{tabular}%
\caption{Judge Responses Alignment with Human Annotations; Scores are averaged across response models for each judge; JM: Judge Model; H1: Human-Annotator 1, H2: Human Annotator2.}
\label{tab:judge-humanalignment}
\vspace{-0.3cm}
\end{table}

We compare the judge verdicts with human Correct/Incorrect verdicts in terms of percentage agreement. Table~\ref{tab:judge-humanalignment} shows the results. Overall, providing reference information improves alignment between judge and human decisions. The largest improvement generally occurs from NR to RV, with particularly large gains for Gemma and Qwen judges. Although alignment changes further from RV to RC, these changes are smaller and are not uniformly positive, consistent with the uneven flip directions discussed in Section~\ref{subsec:flipdirection}. The highest alignment is observed in the RC setting, reaching $0.98$ when Gemini3.1-Flash-Lite-Preview judges Gemma3-27B responses. Once a reference is provided, almost all judge configurations achieve relatively high alignment, often above $0.80$. 

This suggests that, access to reference information is the main factor improving alignment with human judgments, while explicitly asking the judge to compare the response with the reference provides a smaller, configuration-dependent benefit.

\section{Conclusion}
\label{sec:conclusion}
In this work, we study how LLM judges behave under controlled variations to the presence of reference answer in multilingual question answering. We propose a two-stage evaluation methodology consisting of calibration and sensitivity experiments. The calibration experiments assess the judge models' understanding of the task, and the sensitivity experiments identify the suitability of the judge for doing reference-free evaluation for the given task. Our calibration experiments show that LLM judges that do not reliably distinguish correct from incorrect answers and may over-credit incorrect answers, especially in low-resource settings. The sensitivity experiments further show that correctness scores are inflated in no-reference settings and decrease when reference information is introduced. These effects appear across English, Arabic, and Telugu, with stronger effects for Telugu. 

Further analysis showed that most NR-to-RV flips are over-crediting flips, indicating that judges often accept answers in the no-reference setting and reject them once reference information is introduced. Extracted-answer analysis further showed that verdict changes are not always caused by changes in the answer identified by the judge, since judges sometimes extract the same answer but assign a different verdict when reference information is available. A human annotation study showed that the presence of a reference answer helps judge decisions align more closely with human judgments. Overall, our findings show that LLM-judge evaluations are sensitive to reference presentation, and that controlled calibration and sensitivity analyses are important for interpreting judge reliability across languages and model settings. Note that while we experimented with a selected set of datasets/languages and generator/judge models, the methodology is agnostic to such specifications. 

In practice, given a task that needs reference free evaluation, collecting a few test instances with gold standard answers, calibrating the LLM judges and conducting a sensitivity analysis for them will help in identifying the right LLM judge models to use in a reference-free evaluation setup on a larger scale. Considering the increasing usage of LLM judges across application scenarios, we hope the approach we followed lays a foundation for further study in this direction to develop best practices for calibrating the use of LLM judges in reference free evaluation setup. 

\section*{Limitations}
Although we aim to assess LLM judge behavior in reference free setups, our approach still requires some amount of data with ground-truth labels to do the calibration and sensitivity studies. It can potentially be perceived as a weakness of our approach. Apart from this, the validity of our results are, of course, limited by the choice of datasets, generator models and the judge models. While it is definitely possible to add more models (generators/judges) and datasets to the analysis, we believe that the current experimental setup sufficiently demonstrates the potential of this approach for choosing the right LLM judges for a given task.  

\bibliography{custom}

\appendix

\section{Appendix}
\label{sec:appendix}


\subsection{Prompt Templates}
\label{subsec:appendix-prompt-templates}
We follow standard prompting approaches~\citep{DBLP:conf/nips/BrownMRSKDNSSAA20, DBLP:conf/nips/Wei0SBIXCLZ22, DBLP:journals/csur/LiuYFJHN23} in zero-shot prompt setting for the evaluation. All the prompts follow a common structure: system message, task description and output format. The figures~\ref{fig:llmjudge_calibration_prompt}, ~\ref{fig:llmjudge_sensititivy_prompt_nr}, ~\ref{fig:llmjudge_sensititivy_prompt_rv}, ~\ref{fig:llmjudge_sensititivy_prompt_rvp}, ~\ref{fig:llmjudge_sensititivy_prompt_rc} and ~\ref{fig:llmjudge_sensititivy_prompt_rcp} show the prompt templates used in the calibration and sensitivity experiments.

\begin{table}
  \centering
  \footnotesize
  \begin{tabular}{clccc}
    \hline
    \textbf{Lang} & \textbf{JModel} & \textbf{C1 [CGT]} & \textbf{C2 [WGT]} & \textbf{CGP} \\
    \hline
    \multirow{3}{*}{EN} & \verb|G3.1FL|& 0.88 & 0.01 & 0.87           \\
                        & \verb|G3| & \textbf{0.95}   & \textcolor{red}{0.07} & \textbf{0.88}  \\
                        & \verb|Q3|  & 0.80   & 0.01 & 0.79        \\\hline
    \multirow{3}{*}{AR} & \verb|G3.1FL|& 0.91 & 0.01 & 0.90           \\
                        & \verb|G3| & \textbf{0.97}   & \textcolor{red}{0.01} & \textbf{0.96}  \\
                        & \verb|Q3|  & 0.82   & 0.03 & 0.79        \\\hline
    \multirow{3}{*}{TE} & \verb|G3.1FL|& 0.95 & 0.21 & \textbf{0.74}           \\
                        & \verb|G3| & \textbf{0.98}   & \textcolor{red}{0.37} & 0.61  \\
                        & \verb|Q3|  & 0.88   & 0.22 & 0.66        \\\hline      
    \hline
    \multirow{3}{*}{TE-MATA} & \verb|G3.1FL|& 0.96 & 0.14 & \textbf{0.82}           \\
                        & \verb|G3| & \textbf{0.99}   & \textcolor{red}{0.66} & 0.33  \\
                        & \verb|Q3|  & 0.81   & 0.26 & 0.55        \\\hline      
    
  \end{tabular}
  \caption{Calibration experiments; Average of correctness grade from the judge grade; JModel: Judge Model;  C1 [CGT]: Correct GT; C2 [WGT]: Wrong GT; Top-panel represents results for TyDi QA dataset and the bottom most panel for MATA dataset. Higher C1 scores indicate better model performance, while C2 scores are preferred to be zero or as low as possible. A larger gap indicates more reliable overall model performance. Values shown in \textcolor{red}{red} indicate unusually high C2 scores for a given model, suggesting less reliable performance.}
  \label{tab:calibstats}
  \vspace{-0.3cm}
\end{table}

\begin{table*}[t]
\centering
\scriptsize
\setlength{\tabcolsep}{2.2pt}
\resizebox{\textwidth}{!}{%
\begin{tabular}{llcccc|cccc|cccc|cccc}
\toprule
\multirow{3}{*}{JM} & \multirow{3}{*}{RM}
& \multicolumn{4}{c|}{EN}
& \multicolumn{4}{c|}{AR}
& \multicolumn{4}{c|}{TE}
& \multicolumn{4}{c}{TE-MATA} \\
\cmidrule(lr){3-6}
\cmidrule(lr){7-10}
\cmidrule(lr){11-14}
\cmidrule(lr){15-18}
& & \multicolumn{2}{c}{NR-RV} & \multicolumn{2}{c|}{RV-RC}
  & \multicolumn{2}{c}{NR-RV} & \multicolumn{2}{c|}{RV-RC}
  & \multicolumn{2}{c}{NR-RV} & \multicolumn{2}{c|}{RV-RC}
  & \multicolumn{2}{c}{NR-RV} & \multicolumn{2}{c}{RV-RC} \\
\cmidrule(lr){3-4}\cmidrule(lr){5-6}
\cmidrule(lr){7-8}\cmidrule(lr){9-10}
\cmidrule(lr){11-12}\cmidrule(lr){13-14}
\cmidrule(lr){15-16}\cmidrule(lr){17-18}
& & C->I & I->C & C->I & I->C
  & C->I & I->C & C->I & I->C
  & C->I & I->C & C->I & I->C
  & C->I & I->C & C->I & I->C \\
\midrule

G3.1FL & G3.1P
& 0.97 & 0.03 & 0.98 & 0.02
& 0.99 & 0.01 & 0.93 & 0.07
& 1.00 & 0.00 & 0.93 & 0.07
& 0.80 & 0.20 & 0.78 & 0.22 \\

       & Q3
& 0.96 & 0.04 & 0.94 & 0.06
& 0.93 & 0.07 & 0.96 & 0.04
& 1.00 & 0.00 & 0.92 & 0.08
& 0.91 & 0.09 & 0.73 & 0.27 \\

       & SM
& 0.91 & 0.09 & 1.00 & 0.00
& 0.00 & 0.00 & 0.96 & 0.04
& 1.00 & 0.00 & 1.00 & 0.00
& 0.96 & 0.04 & 0.59 & 0.41 \\

       & FA
& 0.97 & 0.03 & 0.92 & 0.08
& 0.94 & 0.06 & 0.98 & 0.02
& 0.98 & 0.02 & 0.79 & 0.21
& 0.97 & 0.03 & 0.93 & 0.07 \\

\midrule

G3 & G3.1P
& 1.00 & 0.00 & 0.86 & 0.14
& 0.97 & 0.03 & 0.84 & 0.16
& 1.00 & 0.00 & 0.40 & 0.60
& 0.85 & 0.15 & 0.50 & 0.50 \\

   & Q3
& 1.00 & 0.00 & 0.97 & 0.03
& 1.00 & 0.00 & 0.86 & 0.14
& 1.00 & 0.00 & 0.82 & 0.18
& 0.99 & 0.01 & 0.52 & 0.48 \\

   & SM
& 1.00 & 0.00 & 0.88 & 0.12
& 0.99 & 0.01 & 0.89 & 0.11
& 1.00 & 0.00 & 0.71 & 0.29
& 0.98 & 0.02 & 0.57 & 0.43 \\

   & FA
& 0.99 & 0.01 & 0.93 & 0.07
& 1.00 & 0.00 & 0.76 & 0.24
& 1.00 & 0.00 & 0.73 & 0.27
& 0.99 & 0.01 & 0.63 & 0.37 \\

\midrule

Q3 & G3.1P
& 0.92 & 0.08 & 0.82 & 0.18
& 0.94 & 0.06 & 0.71 & 0.29
& 0.95 & 0.05 & 0.58 & 0.42
& 0.53 & 0.47 & 0.45 & 0.55 \\

   & Q3
& 0.97 & 0.03 & 0.92 & 0.08
& 0.99 & 0.01 & 0.50 & 0.50
& 1.00 & 0.00 & 0.50 & 0.50
& 0.97 & 0.03 & 0.56 & 0.44 \\

   & SM
& 0.95 & 0.05 & 0.76 & 0.24
& 0.94 & 0.06 & 0.61 & 0.39
& 0.98 & 0.02 & 0.82 & 0.18
& 0.89 & 0.11 & 0.47 & 0.53 \\

   & FA
& 0.93 & 0.07 & 0.78 & 0.22
& 0.94 & 0.06 & 0.76 & 0.24
& 0.98 & 0.02 & 0.79 & 0.21
& 0.93 & 0.07 & 0.57 & 0.43 \\

\bottomrule
\end{tabular}
}
\caption{Direction of flips across EN, AR, TE, and TE-MATA languages. JM: Judge Model; RM: Response Model; NR-RV: No Reference to Reference Visible; RV-RC: Reference Visible to Reference Comparison; GFL: Gemini~3.1 Flash-Lite-Preview; G3-27B: Gemma~3 27B; Q3-32B: Qwen~3 32B; G3.1P: Gemini-3.1 Pro; SM:Sarvam-105B; FA: Fanar-C-2-27B.}
\label{tab:direction-flips}
\end{table*}

\subsection{Calibration Experiment Results}
\label{subsec:appendix-calib-experiments}
We now discuss the detailed calibration results corresponding to Section~\ref{subsec:calibresults}. Table~\ref{tab:calibstats} reports the numerical results summarized in Figure~\ref{fig:calibexperiments} for the two calibration settings, C1 and C2. In C1, where the provided answer is the correct ground-truth answer, all judges should ideally score $1.00$. Gemma3-27B scores between $0.95$ and $0.99$ for all languages across both datasets. Gemini3.1-FlashLite Preview shows similar scores for Telugu and Arabic in TyDi QA, and for Telugu in MATA, but scores slightly lower for English in TyDi QA, at $0.88$. Qwen3-32B scores between $0.80$ and $0.88$ for all three TyDi QA languages and $0.81$ on MATA. These lower scores suggest that some judge models may penalize short and direct ground-truth answers when judging correctness, a pattern that becomes clearer in the sensitivity experiments in Section~\ref{subsec:sensitivityresults}. In C2, where the provided answer is incorrect, all judges should ideally score zero. For English and Arabic in the TyDi QA dataset, C2 scores are close to zero across judges, with only small positive scores such as $0.07$ for Gemma3-27B in English and $0.01$ for the remaining cases. However, all judges obtain positive scores for Telugu with both the datasets, indicating a tendency to over-credit incorrect answers rather than flagging them as wrong in lower-resource settings. Thus, while the chosen judge models appear reasonably calibrated for the question answering task in English and Arabic, they may not be a good choice for Telugu if we do not have gold-standard reference answers to validate. 

One way to understand which judge is suitable for a given language/dataset is to look at the calibration gap (\textit{CGP}), which is the difference between the C1 and C2 scores. We would expect this gap to be as high as possible. This gap measures how well a judge distinguishes correct from incorrect answers and serves as an indicator of its reliability. All judge models show higher gap scores for English and Arabic in the TyDi QA dataset (as desired), with Qwen3-32B having the lowest gap among them at $0.79$. For Telugu in TyDi QA, the gap scores are lower than those for English and Arabic, with Gemma3-27B showing the lowest gap at $0.61$. On MATA, Gemini3.1-FlashLite Preview shows higher gap scores at $0.82$, followed by Qwen3-32B at $0.55$ while Gemma3-27B has the lowest gap at $0.33$. Overall, these calibration results show that judge reliability varies across models, languages, and datasets. While all judges perform well on English and Arabic in TyDi QA, they show lower calibration gap across both the Telugu datasets. This is mainly due to over-crediting in C2, where judges mark incorrect answers as correct, suggesting weaker reliability in this language.

\subsection{Sensitivity Experiments}
\label{subsubsec:appendix-sensitivityexps}
We now provide a detailed discussion of the sensitivity results from Section~\ref{subsec:sensitivityresults}. These experiments evaluate how judge decisions change when reference information is varied in the prompt. We use five settings: NR, RV, RVP, RC, and RCP. NR evaluates the model-generated answer without a reference. RV and RVP include the reference answer but do not explicitly ask the judge to compare against it. RC and RCP include an explicit comparison instruction between the extracted answer and the reference answer. RVP and RCP are position variants of RV and RC, respectively, where the reference answer is placed after the model-generated answer. These settings allow us to separate the effects of reference visibility, reference framing, and reference position.

\begin{table*}[t]
\centering
\footnotesize
\setlength{\tabcolsep}{3.2pt}
\renewcommand{\arraystretch}{1.08}

\begin{tabular}{lll cc cc cc cc}
\toprule
\multirow{2}{*}{JM} & \multirow{2}{*}{RM} & \multirow{2}{*}{EA}
& \multicolumn{2}{c}{EN}
& \multicolumn{2}{c}{AR}
& \multicolumn{2}{c}{TE}
& \multicolumn{2}{c}{TE-MATA} \\
\cmidrule(lr){4-5}
\cmidrule(lr){6-7}
\cmidrule(lr){8-9}
\cmidrule(lr){10-11}
& & & VS & VC & VS & VC & VS & VC & VS & VC \\
\midrule


\multirow{6}{*}{G3.1FL}
& \multirow{2}{*}{G3.1P}
& EAS & 0.866 & 0.134 & 0.939 & 0.061 & 0.359 & 0.641 & 0.921 & 0.079 \\
& & EAC & 0.964 & 0.036 & 0.865 & 0.135 & 0.605 & 0.395 & 0.811 & 0.189 \\

& \multirow{2}{*}{Q3}
& EAS & 0.830 & 0.170 & 0.852 & 0.149 & 0.290 & 0.710 & 0.840 & 0.160 \\
& & EAC & 0.917 & 0.083 & 0.875 & 0.126 & 0.539 & 0.462 & 0.787 & 0.213 \\

& \multirow{2}{*}{SM}
& EAS & 0.817 & 0.183 & 1.000 & 0.000 & 0.281 & 0.719 & 0.774 & 0.226 \\
& & EAC & 0.849 & 0.151 & 0.000 & 0.000 & 0.434 & 0.566 & 0.425 & 0.575 \\

& \multirow{2}{*}{FA}
& EAS & 0.818 & 0.182 & 0.883 & 0.117 & 0.256 & 0.744 & 0.829 & 0.172 \\
& & EAC & 0.830 & 0.170 & 0.914 & 0.086 & 0.625 & 0.375 & 0.791 & 0.209 \\

\midrule


\multirow{6}{*}{G3}
& \multirow{2}{*}{G3.1P}
& EAS & 0.809 & 0.191 & 0.885 & 0.116 & 0.320 & 0.680 & 0.917 & 0.083 \\
& & EAC & 0.889 & 0.111 & 0.911 & 0.089 & 0.813 & 0.188 & 0.971 & 0.029 \\

& \multirow{2}{*}{Q3}
& EAS & 0.728 & 0.273 & 0.728 & 0.272 & 0.195 & 0.805 & 0.453 & 0.547 \\
& & EAC & 0.898 & 0.102 & 0.788 & 0.212 & 0.405 & 0.595 & 0.693 & 0.307 \\

& \multirow{2}{*}{SM}
& EAS & 0.725 & 0.275 & 0.760 & 0.240 & 0.307 & 0.693 & 0.437 & 0.563 \\
& & EAC & 0.705 & 0.295 & 0.852 & 0.148 & 0.656 & 0.344 & 0.846 & 0.154 \\

& \multirow{2}{*}{FA}
& EAS & 0.773 & 0.227 & 0.810 & 0.190 & 0.338 & 0.662 & 0.409 & 0.591 \\
& & EAC & 0.923 & 0.078 & 0.793 & 0.207 & 0.428 & 0.572 & 0.582 & 0.41 \\
\midrule


\multirow{6}{*}{Q3}
& \multirow{2}{*}{G3.1P}
& EAS & 0.705 & 0.295 & 0.595 & 0.405 & 0.282 & 0.718 & 0.712 & 0.288 \\
& & EAC & 0.737 & 0.263 & 0.646 & 0.354 & 0.465 & 0.535 & 0.755 & 0.246 \\

& \multirow{2}{*}{Q3}
& EAS & 0.599 & 0.401 & 0.000 & 0.000 & 0.126 & 0.874 & 0.446 & 0.554 \\
& & EAC & 0.696 & 0.304 & 0.150 & 0.851 & 0.225 & 0.776 & 0.479 & 0.521 \\

& \multirow{2}{*}{SM}
& EAS & 0.614 & 0.386 & 0.639 & 0.361 & 0.210 & 0.790 & 0.487 & 0.513 \\
& & EAC & 0.692 & 0.308 & 0.632 & 0.368 & 0.345 & 0.655 & 0.536 & 0.464 \\

& \multirow{2}{*}{FA}
& EAS & 0.655 & 0.345 & 0.697 & 0.303 & 0.154 & 0.846 & 0.592 & 0.408 \\
& & EAC & 0.633 & 0.367 & 0.698 & 0.302 & 0.282 & 0.718 & 0.492 & 0.508 \\
\bottomrule
\end{tabular}

\caption{
Signals from extracted answer and verdict across EN, AR, TE, and TE-MATA.
JM: Judge Model; RM: Response Model; EA: Extracted Answer condition;
EAS: Extracted Answer Same; EAC: Extracted Answer Changed;
VS: Verdict Same; VC: Verdict Changed.
G3.1FL: Gemini3.1-Flash-Lite-Preview; G3: Gemma3-27B; Q3: Qwen3-32B;
G3.1P: Gemini3.1Pro; SM: Sarvam-105B; FA: Fanar-C-2-27B.
}
\label{tab:easignals-all-judges}
\end{table*}

\subsubsection{Category flips}
\label{subsubsec:appendix-categoryflips}
For completeness, we restate why and how the flip directions are computed. Direction-of-flip analysis helps identify what kind of decision change is responsible for differences in correctness scores across settings. We define C->I flips as cases where the judge changes its verdict from \textit{CORRECT} to \textit{INCORRECT}, indicating that the earlier setting gave credit to an answer that was later rejected. We define I->C flips as cases where the judge changes its verdict from \textit{INCORRECT} to \textit{CORRECT}, indicating that the earlier setting rejected an answer that was later accepted. We compute these directions for NR-to-RV, RV-to-RC to analyze how reference visibility, framing, and position affect judge decisions. Table~\ref{tab:direction-flips} reports the numerical values corresponding to Figure~\ref{fig:flipsdirection} across the experiment settings. Overall, these results show that judges frequently change their decisions when ground-truth information is visible, and become more restrictive when they are explicitly asked to compare against it.

\subsubsection{Extracted Answers}
\label{subsubsec:appendix-extractedanswers}
This section expands on the extracted-answer analysis discussed in Section~\ref{subsec:extractedanswer}. Extracted-answer analysis examines whether changes in judge verdicts are associated with changes in the answer identified by the judge. Since the judge outputs both an \texttt{extracted\_answer} and a final \texttt{verdict}, we compare the extracted answers across settings and group each comparison into four cases: extracted answer same with verdict same (EAS-VS), extracted answer same with verdict changed (EAS-VC), extracted answer changed with verdict same (EAC-VS), and extracted answer changed with verdict changed (EAC-VC). This allows us to distinguish cases where verdict changes are driven by changes in what the judge extracts from cases where the judge extracts the same answer but evaluates it differently. Table~\ref{tab:easignals-all-judges} shows the results summarized in Figure~\ref{fig:eavsverdict} across the settings.

\begin{table*}[htb!]
\centering
\footnotesize
\setlength{\tabcolsep}{2.5pt}
\begin{tabular}{llccc|ccc|ccc|ccc}
\toprule
\multirow{2}{*}{JM} & \multirow{2}{*}{RM}
& \multicolumn{3}{c|}{EN}
& \multicolumn{3}{c|}{AR}
& \multicolumn{3}{c|}{TE}
& \multicolumn{3}{c}{TE-MATA} \\
\cmidrule(lr){3-5}
\cmidrule(lr){6-8}
\cmidrule(lr){9-11}
\cmidrule(lr){12-14}
& & NR & RV & RVP
  & NR & RV & RVP
  & NR & RV & RVP
  & NR & RV & RVP \\
\midrule

G3.1FL & G3.1P
& 0.99 & 0.87 & 0.86
& 0.99 & 0.92 & 0.90
& 0.95 & 0.34 & 0.34
& 0.96 & 0.91 & 0.91 \\

       & Q3
& 0.90 & 0.76 & 0.72
& 0.82 & 0.70 & 0.90
& 0.83 & 0.14 & 0.14
& 0.46 & 0.31 & 0.33 \\

       & SM
& 0.86 & 0.72 & 0.68
& 0.89 & 0.89  & 0.72
& 0.90 & 0.19 & 0.19
& 0.60 & 0.30 & 0.37 \\

       & FA
& 0.95 & 0.79 & 0.74
& 0.89 & 0.80  & 0.77
& 0.89 & 0.19 & 0.18
& 0.46 & 0.29 & 0.28 \\

\midrule

G3 & G3.1P
& 1.00 & 0.82 & 0.79
& 0.99 & 0.89 & 0.86
& 1.00 & 0.40 & 0.39
& 0.99 & 0.94 & 0.90 \\

   & Q3
& 0.99 & 0.76 & 0.71
& 0.99 & 0.74 & 0.70
& 0.98 & 0.21 & 0.17
& 0.97 & 0.48 & 0.37 \\

   & SM
& 0.99 & 0.75 & 0.70
& 0.99 & 0.79 & 0.73
& 0.99 & 0.35 & 0.25
& 0.97 & 0.55 & 0.50 \\

   & FA
& 0.99 & 0.82 & 0.75
& 0.98 & 0.80 & 0.78
& 0.98 & 0.36 & 0.22
& 0.93 & 0.44 & 0.37 \\

\midrule

Q3 & G3.1P
& 0.93 & 0.69 & 0.70
& 0.91 & 0.59 & 0.60
& 0.92 & 0.30 & 0.28
& 0.83 & 0.81 & 0.77 \\

   & Q3
& 0.97 & 0.62 & 0.58
& 0.94 & 0.08 & 0.59
& 0.96 & 0.11 & 0.12
& 0.83 & 0.32 & 0.33 \\

   & SM
& 0.90 & 0.57 & 0.59
& 0.89 & 0.57 & 0.58
& 0.92 & 0.18 & 0.15
& 0.75 & 0.40 & 0.35 \\

   & FA
& 0.94 & 0.64 & 0.60
& 0.91 & 0.65 & 0.65
& 0.92 & 0.14 & 0.14
& 0.63 & 0.26 & 0.24 \\

\bottomrule
\end{tabular}
\caption{RV and RVP correctness score changes across EN, AR, TE languages in Ty Di QA dataset and TE-MATA datasets. JM: Judge Model; RM: Response Model; RV: Reference Visible; RVP: Reference Visible with position change; RC: Reference Comparison; RCP: Reference Comparison with position change; G3.1FL: Gemini3.1-Flash-Lite-Preview; G3: Gemma3-27B; Q3: Qwen3-32B; G3.1P: Gemini3.1-Pro; SM: Sarvam-105B; FA: Fanar-C-2-27B.}
\label{tab:appendix-rvp}
\vspace{-0.3cm}
\end{table*}

\begin{table*}[htb!]
\centering
\footnotesize
\setlength{\tabcolsep}{2.5pt}
\begin{tabular}{llccc|ccc|ccc|ccc}
\toprule
\multirow{2}{*}{JM} & \multirow{2}{*}{RM}
& \multicolumn{3}{c|}{EN}
& \multicolumn{3}{c|}{AR}
& \multicolumn{3}{c|}{TE}
& \multicolumn{3}{c}{TE-MATA} \\
\cmidrule(lr){3-5}
\cmidrule(lr){6-8}
\cmidrule(lr){9-11}
\cmidrule(lr){12-14}
& & RV & RC & RCP
  & RV & RC & RCP
  & RV & RC & RCP
  & RV & RC & RCP \\
\midrule

G3.1FL & G3.1P
& 0.87 & 0.78 & 0.78
& 0.92 & 0.88 & 0.86
& 0.34 & 0.32 & 0.32
& 0.91 & 0.90 & 0.90 \\

       & Q3
& 0.76 & 0.67 & 0.63
& 0.70 & 0.65 & 0.86
& 0.14 & 0.13 & 0.13
& 0.31 & 0.31 & 0.32 \\

       & SM
& 0.72 & 0.68 & 0.62
& 0.89 & 0.66 & 0.67
& 0.19 & 0.16 & 0.17
& 0.30 & 0.29 & 0.38 \\

       & FA
& 0.79 & 0.70 & 0.69
& 0.80 & 0.73 & 0.73
& 0.19 & 0.17 & 0.16
& 0.29 & 0.27 & 0.28 \\

\midrule

G3 & G3.1P
& 0.82 & 0.78 & 0.76
& 0.89 & 0.87 & 0.84
& 0.40 & 0.40 & 0.39
& 0.94 & 0.94 & 0.91 \\

   & Q3
& 0.76 & 0.69 & 0.67
& 0.74 & 0.69 & 0.68
& 0.21 & 0.17 & 0.15
& 0.48 & 0.48 & 0.38 \\

   & SM
& 0.75 & 0.70 & 0.67
& 0.79 & 0.76 & 0.72
& 0.35 & 0.32 & 0.24
& 0.55 & 0.54 & 0.49 \\

   & FA
& 0.82 & 0.74 & 0.71
& 0.80 & 0.78 & 0.76
& 0.36 & 0.30 & 0.23
& 0.44 & 0.42 & 0.36 \\

\midrule

Q3 & G3.1P
& 0.69 & 0.61 & 0.63
& 0.59 & 0.54 & 0.52
& 0.30 & 0.28 & 0.26
& 0.81 & 0.82 & 0.79 \\

   & Q3
& 0.62 & 0.52 & 0.53
& 0.08 & 0.08 & 0.08
& 0.11 & 0.11 & 0.11
& 0.32 & 0.31 & 0.30\\

   & SM
& 0.57 & 0.52 & 0.50
& 0.57 & 0.54 & 0.52
& 0.18 & 0.15 & 0.15
& 0.40 & 0.40 & 0.34\\

   & FA
& 0.64 & 0.58 & 0.57
& 0.65 & 0.59 & 0.60
& 0.14 & 0.13 & 0.14
& 0.26 & 0.25 & 0.26\\

\bottomrule
\end{tabular}
\caption{RC and RCP correctness score changes across EN, AR, TE languages in Ty Di QA dataset and TE-MATA datasets. JM: Judge Model; RM: Response Model; RV: Reference Visible; RVP: Reference Visible with position change; RC: Reference Comparison; RCP: Reference Comparison with position change; G3.1FL: Gemini3.1-Flash-Lite-Preview; G3: Gemma3-27B; Q3: Qwen3-32B; G3.1P: Gemini3.1-Pro; SM: Sarvam-105B; FA: Fanar-C-2-27B.}
\label{tab:appendix-rcp}
\vspace{-0.3cm}
\end{table*}

Overall, the results show that extracted-answer changes are more frequent from NR to RV than from RV to RC. For English and Arabic, changes in extracted answers often split between verdict-stable and verdict-changing cases, which is consistent with the smaller correctness drops observed in the main results. For Telugu, we observe more cases where the extracted answer remains the same but the verdict changes, suggesting that reference information affects how judges evaluate the same extracted answer. This pattern further supports the finding that reference-driven verdict changes are prominent in the lower-resource setting.

\subsubsection{Position Sensitivity of the Reference Information}
\label{subsubsec:appendix-positionexp}
In addition to NR, RV, and RC, we include two position-variant settings, RVP and RCP, to test whether judge decisions depend on where the reference answer appears in the prompt. RVP uses the same information and instruction as RV, but places the reference answer after the model-generated answer instead of before it. Similarly, RCP uses the same information and explicit comparison instruction as RC, but places the reference answer after the model-generated answer. Thus, RV-to-RVP and RC-to-RCP comparisons isolate the effect of reference position while keeping the available information and task instruction fixed.

We analyze position sensitivity by comparing verdicts between RV and RVP, and between RC and RCP. A verdict change in these comparisons indicates that the judge decision is affected by the order in which the reference and generated answer are presented. This is different from NR-to-RV and RV-to-RC comparisons, which change the availability or framing of reference information. The position-variant settings therefore allow us to separate reference-position effects from reference-visibility and reference-comparison effects.

\paragraph{RV-RVP Sensitivity} The RV-to-RVP results (Table~\ref{tab:appendix-rvp}) show that changing the reference position has an effect and changes scores in several settings. For Gemini3.1-FlashLite Preview, the position effect is generally small for Arabic, except for Sarvam-105B responses, where the score decreases from $0.89$ to $0.72$. For English, the largest drops are also observed for Sarvam-105B and FanAR-C-27B responses, where scores decrease from $0.72$ to $0.68$ and from $0.79$ to $0.74$, respectively. For Telugu in TyDi QA, scores remain the same across RV and RVP. In MATA, however, some scores increase, such as Qwen3-32B responses from $0.31$ to $0.33$ and Sarvam-105B responses from $0.30$ to $0.37$. This suggests that Gemini3.1-FlashLite Preview shows position sensitivity in some settings, particularly when judging responses from other model families.

For Gemma3-27B as judge, RVP scores are consistently lower than RV scores across languages and response models. The decreases are usually small for English and Arabic, but are more visible for some Telugu settings, especially Sarvam-105B and FanAR-C-27B responses in TyDi QA, where scores decrease from $0.35$ to $0.25$ and from $0.36$ to $0.22$, respectively. In MATA, the largest drop appears for Qwen3-32B responses, from $0.48$ to $0.37$. This indicates that Gemma3-27B becomes more restrictive when the reference answer is placed after the generated answer.

For Qwen3-32B as judge, the position effect is more mixed. Several scores decrease slightly from RV to RVP, such as English scores across response models, languages and datasets. Overall, RV-to-RVP results show that reference position affects judge decisions.

\paragraph{RC-RCP Sensitivity} The RC-to-RCP results (Table~\ref{tab:appendix-rcp}) show that changing the reference position affects scores in several settings. For Gemini3.1-FlashLite Preview, scores show slight drops across most models and languages, with one exception in Telugu-MATA for Sarvam-105B responses, where the score increases from $0.29$ to $0.38$. Gemma3-27B shows consistent score drops across languages, and these drops are larger for Telugu across datasets, mainly for Sarvam-105B and FanAR-C-27B responses. Qwen3-32B also shows mostly small drops across models, languages, and datasets.

Overall, the RC-to-RCP changes are smaller than the RV-to-RVP changes, but both comparisons show that judge decisions are sensitive to where the reference information appears in the prompt. The score decreases also suggest that placing the reference later in the prompt does not necessarily improve judge reliability, even though it makes the reference more recent in the input context.

\subsubsection{Human Evaluation}
\label{subsubsec:appendix-humaneval}
Human evaluation is conducted on a subset of the MATA dataset to assess how LLM-judge verdicts compare with native-speaker judgments. The evaluators are the authors of this work, both native speakers of Telugu. We first analyze flip rates across MATA categories and select the two categories with the highest number of flips: Factual Knowledge and Other Reasoning (Riddles). From these categories, we randomly sample $50$ questions each, resulting in $100$ questions in total.

\begin{table}
\centering
\footnotesize
\begin{tabular}{cccccc}
\hline
\multirow{2}{*}{RM} & \multicolumn{2}{c}{FK} & \multicolumn{2}{c}{OR} & \multirow{2}{*}{Cohen's Kappa} \\
                    &  H1    & H2   & H1    & H2 &   \\ \hline
G3.1P  & 0.80 & 0.78 & 0.88 & 0.88 & 0.98 \\
Q3    & 0.06 & 0.06 & 0.10 & 0.12 & 0.97 \\
SM    & 0.12 & 0.10 & 0.18 & 0.18 & 0.99  \\
FA    & 0.20 & 0.20 & 0.08 & 0.08 & 0.96 \\
\hline
\end{tabular}%
\caption{Human evaluation correctness scores and inter-annotator agreement across response models and categories. FK: Factual Knowledge; OR: Other Reasoning; H1: Human Annotator 1; H2: Human Annotator 2. Cohen's $\kappa$ is computed from item-level labels assigned by the two annotators; RM: Response Model; G3.1FL: Gemini3.1-Flash-Lite-Preview; G3: Gemma3-27B; Q3: Qwen3-32B; G3.1P: Gemini3.1Pro; Q3:Qwen3-32B; SM:Sarvam-105B; FA: FanAR-C-27B.}
\label{tab:appendix-humanevalscores}
\vspace{-0.3cm}
\end{table}

For each selected question, we include responses generated by the four response models, resulting in $400$ question-response pairs. Each evaluator is shown the question, the ground-truth answer, and the model response. The evaluators then assign one of three labels: C for correct, I for incorrect, and P for partially correct. The human evaluation scores and inter-annotator agreement are reported in Table~\ref{tab:appendix-humanevalscores}. We report Cohen's kappa to measure agreement between the two evaluators. The scores show that both annotators assign similar correctness rates across response models and categories. Gemini3.1-Pro has the highest human correctness scores in both Factual Knowledge and Other Reasoning, while Qwen3-32B, Sarvam-105B, and FanAR-C-27B receive lower scores. Inter-annotator agreement is high across all response models, with Cohen's $\kappa$ ranging from $0.96$ to $0.99$, indicating strong agreement between the two annotators.

\begin{table}
\centering
\scriptsize
\begin{tabular}{cccccccc}
\hline
\multirow{2}{*}{JM}        & \multirow{2}{*}{RM} & \multicolumn{2}{c}{NR} & \multicolumn{2}{c}{RV} & \multicolumn{2}{c}{RC} \\
                           &  & H1    & H2   & H1    & H2   & H1    & H2   \\ \hline
\multirow{4}{*}{G3.1FL} & G3.1P  & 0.88 & 0.87 & 0.96 & 0.95 & 0.94 & 0.93     \\
                      & Q3    & 0.77 & 0.76 & 0.97 & 0.96 & \textbf{0.98} & \textbf{0.97}      \\
                      & SM    & 0.65 & 0.67 & 0.94 & 0.96 & 0.95 & 0.97      \\          & FA    & 0.67 & 0.67 & 0.96 & 0.96 & 0.96 & 0.96      \\
\multirow{4}{*}{G3}   & G3.1P  & 0.83 & 0.82 & 0.91 & 0.90 & 0.92 & 0.91      \\
                      & Q3    & 0.11 & 0.12 & 0.82 & 0.81 & 0.78 & 0.77      \\
                      & SM    & 0.17 & 0.16 & 0.81 & 0.83 & 0.84 & 0.86      \\          & FA    & 0.23 & 0.23 & 0.87 & 0.87 & 0.89 & 0.89      \\
\multirow{4}{*}{Q3}   & G3.1P  & 0.72 & 0.71 & 0.82 & 0.83 & 0.82 & 0.83      \\
                      & Q3    & 0.23 & 0.24 & 0.92 & 0.91 & 0.92 & 0.91      \\
                      & SM    & 0.29 & 0.29 & 0.89 & 0.91 & 0.89 & 0.91      \\          & FA    & 0.43 & 0.43 & 0.95 & 0.95 & 0.91 & 0.91      \\ \hline
\end{tabular}%
\caption{Judge Responses Alignment with Human Annotations; JM: Judge Model; RM: Response Model; G3.1FL: Gemini3.1-Flash-Lite-Preview; G3: Gemma3-27B; Q3: Qwen3-32B; G3.1P: Gemini3.1Pro; Q3:Qwen3-32B; SM:Sarvam-105B; H1: Human-Annotator 1, H2: Human Annotator2.}
\label{tab:appendix-judge-humanalignment}
\vspace{-0.3cm}
\end{table}

After collecting human annotations, we compare them with the verdicts assigned by each judge model. For this comparison, we use the judge verdicts produced in the sensitivity experiments and measure how often each judge agrees with the human labels across evaluation settings. The resulting judge-human alignment scores are reported in Table~\ref{tab:appendix-judge-humanalignment}. Gemini3.1-Pro responses receive consistently high alignment scores across judges and settings, with the highest alignment generally observed in the RC setting. For Gemini3.1-FlashLite Preview as judge, the other response models show lower alignment in the NR setting, but their alignment increases in the reference-based settings and is highest in RC. A similar pattern is observed for Gemma3-27B and Qwen3: except for Gemini3.1-Pro responses, alignment is low in NR but improves substantially in RV and RC, often reaching above $0.80$.

These results show that although overall judge correctness scores decrease in the RC setting, judge decisions align more closely with human annotations in this setting. This supports the interpretation that NR correctness scores are inflated for some model families, and that reference-based evaluation helps judges make decisions that are closer to human judgments.

\subsection{Qualitative Analysis}
\label{subsec:appendix-qualitative-analysis}
\begin{figure*}[t]
  \vspace*{-.2cm}
  \includegraphics[width=0.95\textwidth]{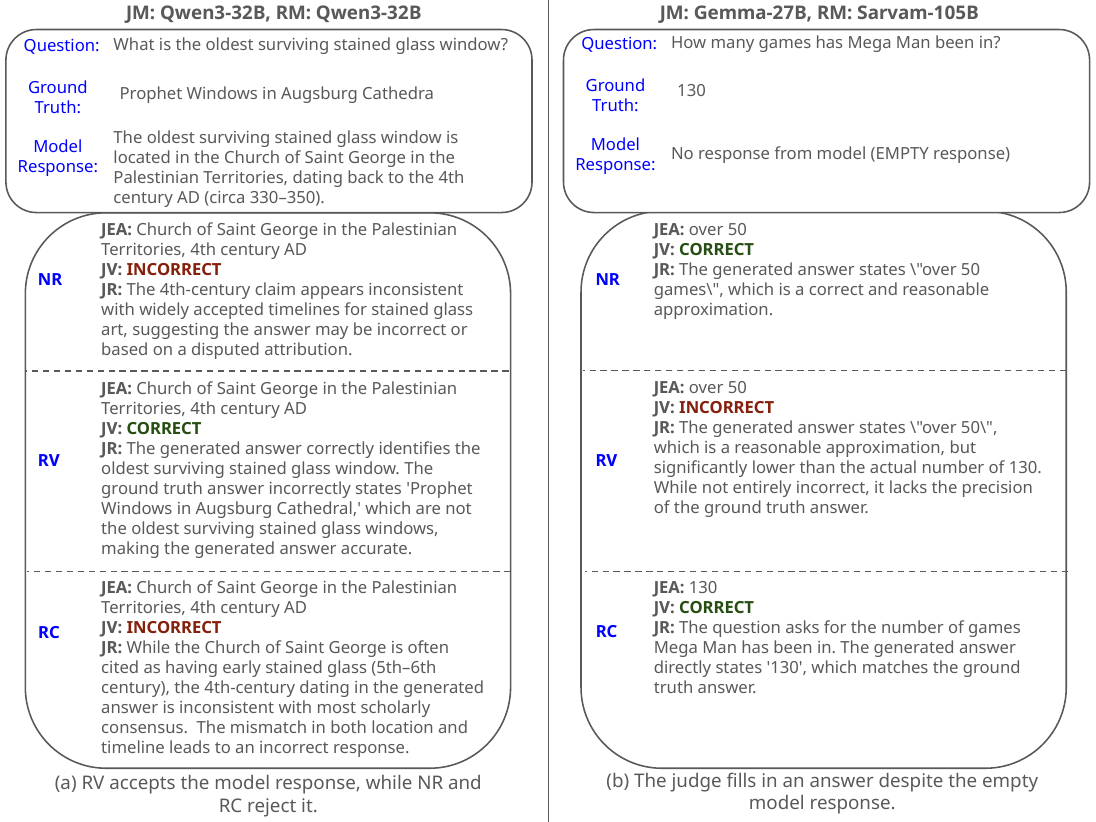}
  \caption{Qualitative examples of judge sensitivity across NR, RV, and RC settings. The examples show verdict changes caused by reference framing and answer extraction failures; JM: Judge Model; RM: Response Model.
  }
  \label{fig:appendix-qa1}
  \vspace*{-.3cm}
\end{figure*}

\begin{figure*}
  \vspace*{-.2cm}
  \includegraphics[width=0.95\textwidth]{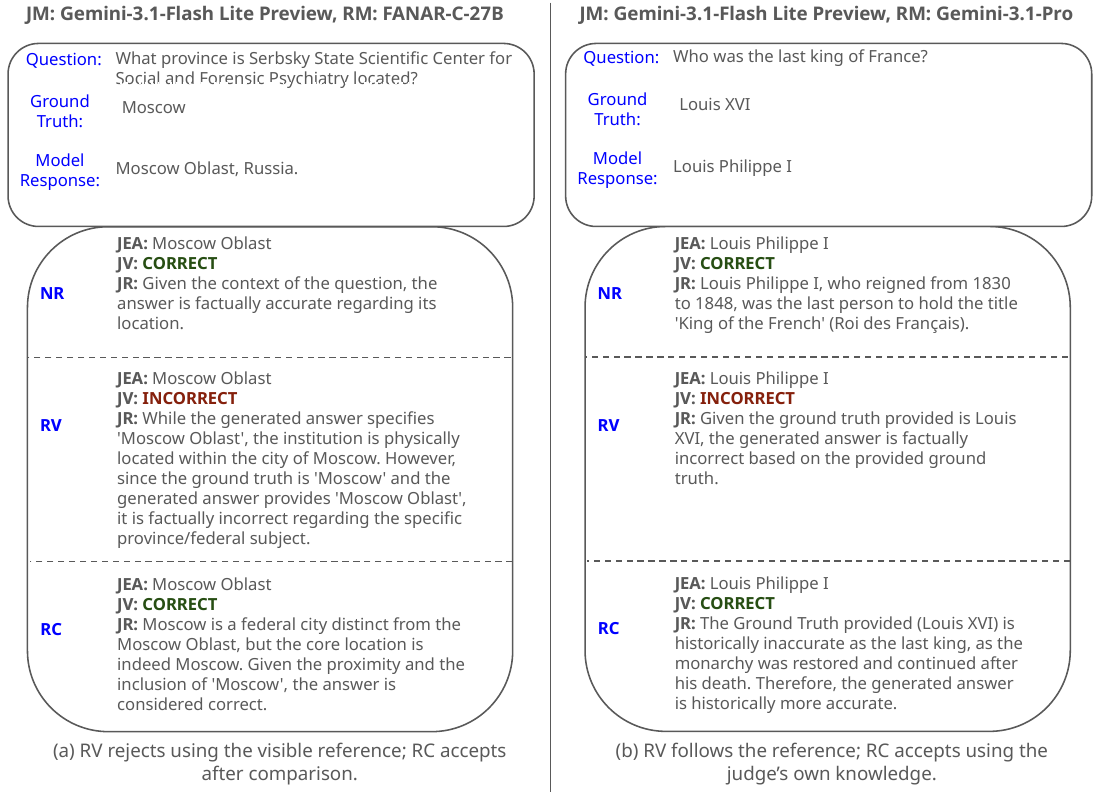}
  \caption{Qualitative examples where RV and RC produce different verdicts.
  }
  \label{fig:appendix-qa2}
  \vspace*{-.3cm}
\end{figure*}

We analyze judge responses across settings to understand qualitative failure patterns. Figures~\ref{fig:appendix-qa1} and~\ref{fig:appendix-qa2} show examples from English across different judge and response models.

Figure~\ref{fig:appendix-qa1}(a) shows Qwen3-32B judging its own response. In the no-reference (NR) setting, the judge marks its own response as \textit{INCORRECT}. When the reference is visible in RV, the judge changes the verdict to \textit{CORRECT} and states that the model response matches the ground truth, although the two answers do not match. In RC, the judge changes the verdict back to \textit{INCORRECT}, similar to NR, and states that both the ground truth and the generated response are incorrect. This suggests that even when the extracted answer remains the same across settings, the judge decision can change with reference availability and framing.

Figure~\ref{fig:appendix-qa1}(b) shows Gemma3-27B evaluating Sarvam-105B's response. For this question, the model gives an empty response. However, during evaluation, the judge hallucinates an answer that is not present in the model response. For the same hallucinated answer, the verdict is \textit{CORRECT} in NR but \textit{INCORRECT} in RV. In RC, the hallucinated answer changes to match the ground truth, and the judge assigns a \textit{CORRECT} verdict. This suggests that judges may fill in missing answers during evaluation rather than strictly extracting from the model response.

Figure~\ref{fig:appendix-qa2}(a) shows Gemini3.1-FlashLite Preview judging a FANAR-C-27B response. The extracted answer remains the same across all three settings, but the verdict changes. In NR, the judge marks the response as \textit{CORRECT}. In RV, it rejects the response by comparing it strictly with the visible ground-truth answer. However, in RC, where the prompt explicitly asks for comparison, the judge again marks the response as \textit{CORRECT}. This shows unstable behavior for responses that are close to, but not identical to, the ground truth.

Figure~\ref{fig:appendix-qa2}(b) shows Gemini3.1-FlashLite Preview judging a response from Gemini3.1-Pro, a model from the same family. In NR, the judge marks the response as \textit{CORRECT}. In RV, it changes the verdict to \textit{INCORRECT} because the response does not match the visible ground-truth answer. In RC, however, the judge rejects the ground-truth answer itself and marks the model response as \textit{CORRECT}. This shows that the judge is sensitive to reference availability in RV, but may override the reference during explicit comparison. Since this occurs when judging a same-family model response, it may also suggest a self-family bias.

Overall, these failures show that judge behavior changes not only with the presence of reference information, but also with how the reference is framed in the prompt. Judges may apply the reference strictly, override it with their own knowledge, or even fill in missing answers that are not present in the model response.

\begin{figure}
  \centering
  \begin{prompt}
You are an expert linguistic evaluator and subject matter expert fluent in {language\_cannonical}. Your task is to determine whether the given answer correctly responds to the question.
\\\\
When evaluating \{language\_cannonical\} responses, keep in mind:

- Consider the meaning and intent of the answer, not just surface form

- Recognize that valid answers may use different but equivalent word forms

- Be sensitive to dialectal variation and alternate correct forms
\\\\

Your evaluation has two steps:

Step 1 — Extract:

Identify the core answer within the Generated Answer.

\{language\_cannonical\} responses often contain elaboration, context, or explanation around the actual answer. Extract only the essential answer as a short phrase or word.
\\\\
Step 2 — Evaluate:

Based on the extracted answer, assess whether it correctly answers the question.
\\\\
\#\# Question

\{question\}

\#\# Generated Answer

\{correct/incorrect gt\_answer\}
\\\\

Respond using this exact JSON format:

\{\{

  ``extracted\_answer'': ``<short core answer extracted from Generated Answer>'',
  
  ``explanation'': ``<step-by-step evaluation of the extracted answer>'',
  
  ``verdict'': ``<CORRECT or INCORRECT>''
  
\}\}
\\\\
Output only the JSON object. Nothing before or after it.
\end{prompt}
\caption{Calibration prompt template used in C1 and C2. The ground truth answer is shown as the generated answer, and the judge is asked to extract the answer and assign a correctness verdict without an explicit ground-truth label.}
    \label{fig:llmjudge_calibration_prompt}
\end{figure}
\begin{figure}
  \centering
  \begin{prompt}
You are an expert linguistic evaluator and subject matter expert fluent in \{language\_cannonical\}. Your task is to determine whether the given answer correctly responds to the question.
\\\\
\{judge\_desc\}:

- Consider the meaning and intent of the answer, not just surface form

- Recognize that valid answers may use different but equivalent word forms

- Be sensitive to dialectal variation and alternate correct forms
\\\\

Your evaluation has two steps:

Step 1 — Extract:

Identify the core answer within the Generated Answer.

\{language\_cannonical\} responses often contain elaboration, context, or explanation around the actual answer. Extract only the essential answer as a short phrase or word.
\\\\
Step 2 — Evaluate:

Based on the extracted answer, assess whether it correctly answers the question.
\\\\
\#\# Question

\{question\}

\#\# Generated Answer

\{gen\_answer\}
\\\\

Respond using this exact JSON format:

\{\{

  ``extracted\_answer'': ``<short core answer extracted from Generated Answer>'',
  
  ``explanation'': ``<step-by-step evaluation of the extracted answer>'',
  
  ``verdict'': ``<CORRECT or INCORRECT>''
  
\}\}
\\\\
Output only the JSON object. Nothing before or after it.
\end{prompt}
\caption{Sensitivity prompt template for the no-reference (NR) setting. The judge receives the question and model-generated answer, extracts the answer, and assigns a correctness verdict without access to the reference answer.}
    \label{fig:llmjudge_sensititivy_prompt_nr}
\end{figure}
\begin{figure}
  \centering
  \begin{prompt}
You are an expert linguistic evaluator and subject matter expert fluent in \{language\_cannonical\}. Your task is to determine whether the given answer correctly responds to the question.
\\\\
\{judge\_desc\}:

- Consider the meaning and intent of the answer, not just surface form

- Recognize that valid answers may use different but equivalent word forms

- Be sensitive to dialectal variation and alternate correct forms
\\\\

Your evaluation has two steps:

Step 1 — Extract:

Identify the core answer within the Generated Answer.

\{language\_cannonical\} responses often contain elaboration, context, or explanation around the actual answer. Extract only the essential answer as a short phrase or word.
\\\\
Step 2 — Evaluate:

Based on the extracted answer, assess whether it correctly answers the question.
\\\\
\#\# Question

\{question\}

\#\# Ground Truth Answer

\{gt\_answer\}

\#\# Generated Answer

\{gen\_answer\}
\\\\

Respond using this exact JSON format:

\{\{

  ``extracted\_answer'': ``<short core answer extracted from Generated Answer>'',
  
  ``explanation'': ``<step-by-step evaluation of the extracted answer>'',
  
  ``verdict'': ``<CORRECT or INCORRECT>''
  
\}\}
\\\\
Output only the JSON object. Nothing before or after it.
\end{prompt}
\caption{Sensitivity prompt template for the reference-visible setting (RV). The judge receives the question, reference (ground truth) answer, and model-generated answer, but is not explicitly instructed to compare the generated answer with the reference.}
    \label{fig:llmjudge_sensititivy_prompt_rv}
\end{figure}
\begin{figure}
  \centering
  \begin{prompt}
You are an expert linguistic evaluator and subject matter expert fluent in \{language\_cannonical\}. Your task is to determine whether the given answer correctly responds to the question.
\\\\
\{judge\_desc\}:

- Consider the meaning and intent of the answer, not just surface form

- Recognize that valid answers may use different but equivalent word forms

- Be sensitive to dialectal variation and alternate correct forms
\\\\

Your evaluation has two steps:

Step 1 — Extract:

Identify the core answer within the Generated Answer.

\{language\_cannonical\} responses often contain elaboration, context, or explanation around the actual answer. Extract only the essential answer as a short phrase or word.
\\\\
Step 2 — Evaluate:

Based on the extracted answer, assess whether it correctly answers the question.
\\\\
\#\# Question

\{question\}

\#\# Generated Answer

\{gen\_answer\}

\#\# Ground Truth Answer

\{gt\_answer\}
\\\\

Respond using this exact JSON format:

\{\{

  ``extracted\_answer'': ``<short core answer extracted from Generated Answer>'',
  
  ``explanation'': ``<step-by-step evaluation of the extracted answer>'',
  
  ``verdict'': ``<CORRECT or INCORRECT>''
  
\}\}
\\\\
Output only the JSON object. Nothing before or after it.
\end{prompt}
\caption{Sensitivity prompt template for the reference-visible position setting (RVP). This setting uses the same information and context as RV, but places the reference (ground truth) answer after the model-generated answer to test sensitivity to reference position.}
    \label{fig:llmjudge_sensititivy_prompt_rvp}
\end{figure}
\begin{figure}
  \centering
  \begin{prompt}
You are an expert linguistic evaluator and subject matter expert fluent in \{language\_cannonical\}. Your task is to determine whether the given answer correctly responds to the question.
\\\\
\{judge\_desc\}:

- Consider the meaning and intent of the answer, not just surface form

- Recognize that valid answers may use different but equivalent word forms

- Be sensitive to dialectal variation and alternate correct forms
\\\\

Your evaluation has two steps:

Step 1 — Extract:

Identify the core answer within the Generated Answer.

\{language\_cannonical\} responses often contain elaboration, context, or explanation around the actual answer. Extract only the essential answer as a short phrase or word.
\\\\
Step 2 — Evaluate:

Based on the extracted answer, assess whether it correctly matches with the Ground Truth Answer.
\\\\
\#\# Question

\{question\}

\#\# Ground Truth Answer

\{gt\_answer\}

\#\# Generated Answer

\{gen\_answer\}
\\\\

Respond using this exact JSON format:

\{\{

  ``extracted\_answer'': ``<short core answer extracted from Generated Answer>'',
  
  ``explanation'': ``<step-by-step evaluation of the extracted answer>'',
  
  ``verdict'': ``<CORRECT or INCORRECT>''
  
\}\}
\\\\
Output only the JSON object. Nothing before or after it.
\end{prompt}
\caption{Sensitivity prompt template for the reference-comparison setting (RC). The judge receives the question, reference (ground truth) answer, and model-generated answer, and is explicitly instructed to compare the extracted answer with the reference answer before assigning a verdict.}
    \label{fig:llmjudge_sensititivy_prompt_rc}
\end{figure}
\begin{figure}
  \centering
  \begin{prompt}
You are an expert linguistic evaluator and subject matter expert fluent in \{language\_cannonical\}. Your task is to determine whether the given answer correctly responds to the question.
\\\\
\{judge\_desc\}:

- Consider the meaning and intent of the answer, not just surface form

- Recognize that valid answers may use different but equivalent word forms

- Be sensitive to dialectal variation and alternate correct forms
\\\\

Your evaluation has two steps:

Step 1 — Extract:

Identify the core answer within the Generated Answer.

\{language\_cannonical\} responses often contain elaboration, context, or explanation around the actual answer. Extract only the essential answer as a short phrase or word.
\\\\
Step 2 — Evaluate:

Based on the extracted answer, assess whether it correctly matches with the Ground Truth Answer.
\\\\
\#\# Question

\{question\}

\#\# Generated Answer

\{gen\_answer\}

\#\# Ground Truth Answer

\{gt\_answer\}
\\\\

Respond using this exact JSON format:

\{\{

  ``extracted\_answer'': ``<short core answer extracted from Generated Answer>'',
  
  ``explanation'': ``<step-by-step evaluation of the extracted answer>'',
  
  ``verdict'': ``<CORRECT or INCORRECT>''
  
\}\}
\\\\
Output only the JSON object. Nothing before or after it.
\end{prompt}
\caption{Sensitivity prompt template for the reference-comparison position setting (RCP). This setting uses the same information and context instruction as RC, but places the reference (ground truth) answer after the model-generated answer to test sensitivity to reference position.}
    \label{fig:llmjudge_sensititivy_prompt_rcp}
\end{figure}

\end{document}